\documentclass[acmsmall]{acmart}

\AtBeginDocument{%
  }

\setcopyright{acmcopyright}
\copyrightyear{2022}
\acmYear{2022}
\acmDOI{XXXXXXX.XXXXXXX}

\acmJournal{CSUR}

\usepackage{algorithm}
\usepackage{algorithmic}
\usepackage{graphicx}
\usepackage{textcomp}
\usepackage{xcolor}
\usepackage{multirow}
\usepackage{xspace}
\usepackage{listings}
\usepackage{subfigure}
\usepackage{bm}
\usepackage{pifont}
\usepackage{CJKutf8}

\definecolor{mycolor1}{RGB}{0,112,192}

\newcommand{\lhzblue}[1]{\textcolor{mycolor1}{#1}}

\newcommand{\blue}[1]{{#1}}

\newcommand{\tabincell}[2]{\begin{tabular}{@{}#1@{}}#2\end{tabular}}

\newcommand{\numsys}{28\xspace}

\newcommand{\cmark}{\ding{51}}%
\newcommand{\xmark}{\ding{55}}%

\newcommand{\showDOI}[1]{\unskip}

\definecolor{dkgreen}{rgb}{0,0.6,0}
\definecolor{gray}{rgb}{0.5,0.5,0.5}
\definecolor{mauve}{rgb}{0.58,0,0.82}

\lstdefinelanguage{c++}{
    morekeywords={asm,do,if,return,typedef,auto,
    double,inline,short,typeid,bool,dynamic_cast,
    int,signed,typename,break,else,long,sizeof,
    union,case,enum,mutable,static,unsigned,catch,
    explicit,namespace,static_cast,using,char,
    export,new,struct,virtual,class,extern,
    operator,switch,void,const,false,private,
    template,volatile,const_cast,float,protected,
    this,wchar_t,continue,for,public,throw,while,
    default,friend,register,true,delete,goto,
    reinterpret_cast,try},
    otherkeywords={=>,<-,<\%,<:,>:,\#,@},
    sensitive=true,
    showspaces=false,
    morecomment=[l]{//},
    morecomment=[n]{/*}{*/},
    morestring=[b]",
    morestring=[b]',
    morestring=[b]"""
}

\begin{document}

\title{Distributed Graph Neural Network Training: A Survey}

%
\author{Yingxia Shao}
\email{shaoyx@bupt.edu.cn}
\affiliation{%
  \institution{Beijing University of Posts and Telecommunications}
  \country{China}
}

\author{Hongzheng Li}
\email{Ethan_Lee@bupt.edu.cn}
\affiliation{%
  \institution{Beijing University of Posts and Telecommunications}
  \country{China} 
}

\author{Xizhi Gu}
\email{guxizhi@bupt.edu.cn}
\affiliation{%
 \institution{Beijing University of Posts and Telecommunications}
  \country{China}
}

\author{Hongbo Yin}
\email{yinhbo@bupt.edu.cn}
\affiliation{%
 \institution{Beijing University of Posts and Telecommunications}
  \country{China}
}

\author{Yawen Li}
\email{lywbupt@126.com}
\affiliation{%
 \institution{Beijing University of Posts and Telecommunications}
  \country{China}
 }

\author{Xupeng Miao}
\email{xupeng@cmu.edu}
\affiliation{%
 \institution{Carnegie Mellon University}
  \country{USA}
 }

 \author{Wentao Zhang}
\email{wentao.zhang@mila.quebec}
\affiliation{%
 \institution{{Mila - Québec AI Institute, HEC Montréal}}
  \country{Canada}
 }

\author{Bin Cui}
\email{bin.cui@pku.edu.cn}
\affiliation{%
 \institution{Peking University}
  \country{China}
  }

\author{Lei Chen}
\email{leichen@cse.ust.hk}
\affiliation{%
  \institution{The Hong Kong University of Science and Technology (Guangzhou)}
  \country{China}
}



\renewcommand{\shortauthors}{Shao et al.}

\begin{abstract}
\blue{Graph neural networks (GNNs) are a type of deep learning models that are trained on graphs and have been successfully applied in various domains.}
Despite the effectiveness of GNNs, it is still challenging for GNNs to efficiently scale to large graphs. 
As a remedy, distributed computing becomes a promising solution of training large-scale GNNs, since it is able to provide abundant computing resources. 
However, the dependency of graph structure increases the difficulty of achieving high-efficiency distributed GNN training, which suffers from the massive communication and workload imbalance. 
In recent years, many efforts have been made on distributed GNN training, and an array of training algorithms and systems have been proposed. 
Yet, there is a lack of systematic review on the optimization techniques for the distributed execution of GNN training.
In this survey, we analyze three major challenges in distributed GNN training that are massive feature communication, the loss of model accuracy and workload imbalance. 
Then we introduce a new taxonomy for the optimization techniques in distributed GNN training that address the above challenges. The new taxonomy classifies existing techniques into four categories that are GNN data partition, GNN batch generation, GNN execution model, and GNN communication protocol. 
We carefully discuss the techniques in each category. 
In the end, we summarize existing distributed GNN systems for multi-GPUs, GPU-clusters and CPU-clusters, respectively, and give a discussion about the future direction on distributed GNN training.
\end{abstract}

\maketitle

\section{Introduction}
\label{sec:intro}
GNNs are powerful tools to handle problems modeled by the graph and have been widely adopted in various applications, including social networks (e.g., social spammer detection~\cite{gnn_socialnetwork, gnn_spammer}, social network analysis~\cite{gnn_sna}), bio-informatics (e.g., protein interface prediction~\cite{gnn_bio}, disease–gene association~\cite{gnn_bio1}), drug discovery~\cite{app_protien, app_drug1}, traffic forecasting~\cite{gnn_traffic}, health care~\cite{gnn_health1,gnn_health2}, recommendation~\cite{gnn_social_rec, gnn_rec, app_rec3, app_news_rec}, natural language process~\cite{gnn_text, gnn_text1} and others~\cite{app_physics, app_chemical, app_combop, app_traffic,app_bio}. By integrating the information of graph structure into the deep learning models, GNNs can achieve significantly better results than traditional machine learning and data mining methods.

A GNN model generally contains multi graph convolutional layers, where each vertex aggregates the latest states of its neighbors, updates the state of the vertex, and applies neural network to the updated state of the vertex. Taking the traditional graph convolutional network (GCN) as an example, in each layer, a vertex uses a sum function to aggregate the neighbor states and its own state, then applies a single-layer MLP to transform the new state. Such procedures are repeated $L$ times if the number of layers is $L$. The vertex states generated in the $L$th layer are used by the downstream tasks, like node classification, link prediction, and so on. In the past years, many research works have made remarkable progress in the design of graph neural network models. Prominent models include GCN~\cite{gcn}, GraphSAGE~\cite{graphsage_nips_2017}, GAT~\cite{gat}, GIN~\cite{gin}, and many other application-specific GNN models~\cite{gnnmodel_city, pmlr-v97-zhang19e}. Up to date, there are tens of surveys reviewing the GNN models~\cite{9039675, PMID:32217482, ZHOU202057, survey_2022}. 
On the other hand, to efficiently develop different GNN models, many GNN-oriented frameworks are proposed based on various deep learning libraries~\cite{Liu2021a,Grattarola2021,Battaglia2018,pygeometric_iclr_2019,cogdl_arxiv21,dgl_arxiv_2019}. Many new optimizations are proposed to speed up GNN training, including GNN computation kernels \cite{pcgcn_ipdps20,fusegnn_iccad20,gespmm_sc20,fusedmm_ipdps21,Zhang2022,tlpgnn_2022,UBG_ppopp_2021}, efficient programming models~\cite{graphiler_mlsys22,seastar_eurosys21,featgraph_SC20}, and full utilization of new hardware~\cite{rubic_2022,graphact_pfga20,gnnear_arxiv21,graphite_ISCA22}.
However, these frameworks and optimizations mainly focus on training GNN on a single machine, while not paying much attention to the scalability of input graphs.

Nowadays, large-scale graph neural networks~\cite{ijcai2022-772, joshi2022dlforrouting} become a hot topic because of the prevalence of massive large graph data. It is common to have graphs with billions of vertices and trillions of edges, like the social network in Sina Weibo, WeChat, Twitter and Meta. 
However, most of the existing GNN models are only tested on small graph data sets, and it is impossible or inefficient to process large graph data sets~\cite{ogb_benchmark}. This is because GNN models are complex and require massive computation resources when handling large graphs. 
A line of works achieve large-scale GNNs by designing scalable GNN models. They use simplification~\cite{SGC_icml_2019, lightgcn_sigir_2020, sign_icml_grl2020}, quantization~\cite{degreequant_iclr_2021,binarygcn_CVPR_2021,Binary_CVPR_2021,2022_Neurocomputing_EPQuant,2022_ICLR_EXACT,2022_PPoPP_QGTC,2021_EuroMLSys_LLPGNN,2021_WWW_BinariedGNN,sgquant_ictai_2020}, sampling~\cite{L2GCN_CVPR_2020, graphsaint_iclr_2020, clusterGCN_kdd_2019} and distillation~\cite{tinyGCN_kdd_2020,rdd_sigmod_2020,ijcai2021-320} to design efficient models.
Another line of works adopt distributed computing to the GNN training, a.k.a, distributed GNN training. Because when handling large graphs, the limited memory and computing resource of a single device (e.g, GPU) become the bottleneck of large-scale GNN training, and distributed computing provides more computing resources (e.g., multi-GPUs, CPU clusters, etc.) to improve the training efficiency.
\blue{
However, previous systems targeting at distributed graph processing~\cite{pregel_sigmod10, dist_graphlab_vldb12} and distributed deep learning~\cite{pytorch_dist_vldb20, tensorflow} separately. 
The graph processing systems do not consider the acceleration of neural network operations, while the deep learning systems lack the ability of processing graph data. 
Therefore, many efforts have been made in designing efficient distributed GNN training frameworks and systems~\cite{wan2022pipegcn, BNSGCN, aligraph_vldb_2019, distdgl_ai3_2020, roc_mlsys_2020}. 
}

\blue{In this survey, we focus on the works and specific techniques proposed for distributed GNN training.} Distributed GNN training divides the whole workload of model training among a set of workers, and all the workers process the workload in parallel. However, due to the data dependency in GNNs, it is non-trivial to apply existing distributed machine learning methods~\cite{ml_survey, ml_survey2} to GNNs, and many new techniques for optimizing the distributed GNN training pipeline are proposed. Although there are a lot of surveys~\cite{9039675, PMID:32217482, ZHOU202057} about GNNs, to the best of our knowledge, little effort has been made to systematically review the techniques for distributed GNN training. Recently, Besta et al.~\cite{Besta2022} only reviewed the parallel computing paradigm of GNN, Abadal~\cite{survey_hardware} surveyed GNN computing from algorithms to hardware accelerators, and 
\blue{Vatter et al.~\cite{distgnn-evolution-comp-survey23} provided a comprehensive overview of the evolution of the distributed systems for scalable GNN training.}

\blue{
To clearly organize the techniques for distributed GNN training, we introduce a general distributed GNN training pipeline which consists of three stages -- data partition, batch generation, and
GNN {model training}.
These stages involve GNN-specific execution logic that includes graph processing and graph aggregation.}  
In the context of this general distributed GNN training pipeline, we discuss three main challenges of distributed GNN training which are caused by the data dependency in graph data and \blue{require new techniques specifically designed for distributed GNN training.}
{To help readers understand various optimization techniques that address the above challenges better, we introduce a new taxonomy that classifies the techniques into four orthogonal categories that are GNN data partition, GNN batch generation, GNN execution model and GNN communication protocol. 
This taxonomy not only covers the optimization techniques used in both mini-batch distributed GNN training and full-graph distributed GNN training, but also discusses the techniques from graph processing to model execution.}  
We carefully review the existing techniques in each category followed by describing {\numsys} representative distributed GNN systems and frameworks either from industry or academia. Finally, we briefly discuss the future directions of distributed GNN training.

The contributions of this survey are as follows:
\begin{itemize}
	\item This is the first survey focusing on the optimization techniques for efficient distributed GNN training, and it helps researchers quickly understand the landscape of distributed GNN training.
	\item {We introduce a new taxonomy of distributed GNN training techniques by considering the life-cycle of end-to-end distributed GNN training. In high-level, the new taxonomy consists of four orthogonal categories that are GNN data partition, GNN batch generation, GNN execution model and GNN communication protocol.}
	\item  We provide a detailed and comprehensive technical summary of each category in our new taxonomy.
	\item We review \numsys representative distributed GNN training systems and frameworks from industry to academia.
	\item We make a discussion about the future directions of the distributed GNN training.
\end{itemize}

\blue{
\textbf{Survey organization}. In Section~\ref{sec:preliminary}, we introduce the background of GNNs, and discuss the differences between GNN training and traditional neural network training. In Section~\ref{sec:pipeline-challenges}, we present the distributed GNN training pipeline, highlight the specific challenges faced by distributed GNN training and introduce our taxonomy of the techniques used in distributed GNN training. On the basis of the taxonomy, we make detailed discussion about the techniques of distributed GNN training, including data partition (Section~\ref{sec:data-partition}), batch generation (Section~\ref{sec:batch-generation}), execution model (Section~\ref{sec:exec-model}), and communication protocol (Section~\ref{sec:communication-protocol}). In Section~\ref{sec:system}, we discuss various existing distributed GNN training systems. In the end, we unveil some promising future directions for distributed GNN training and make a conclusion.
}

\section{Preliminaries of Graph Neural Networks}
\label{sec:preliminary}
\textbf{Graph \& Graph partition}. 
A graph $G=(V, E)$ consists of a vertex set $V$ and an edge set $E$. 
The set of neighbors of a vertex $v\in V$ is denoted by $N(v)$ and the degree of $v$ is denoted by {$d=|N(v)|$}.
In directed graph, the set of incoming and outgoing neighbors of a vertex $v\in V$ is denoted by $N_{in}(v)$ and $N_{out}(v)$, respectively, and the corresponding degree is denoted by $d_{in}=|N_{in}(v)|$ and $d_{out}=|N_{out}(v)|$.
An edge $e \in E$ is represented by $(u,v)$ where $u$, $v \in V$. 
The adjacent matrix of $G$ is represented by $\bm{A}$, where an entry $a_{uv}$ of $\bm{A}$ equals 1 when $(u,v)\in E$, otherwise $a_{uv}=0$. 
Each vertex $v$ in a graph has an initial feature vector $\bm{x_v} \in \mathbb{R}^{D}$, where $D$ is the dimension size of feature vectors, and the feature matrix of the graph is denoted by $\bm{X}$. 

A graph can be partitioned in distributed settings. 
For an edge-cut graph partition, a graph $G=(V, E)$ is divided into $P$ partitions {$\{G_i=(V_i, E_i)\}$, $1\le i,j\le P$, which satisfy $V=\cup V_i, V_i \cap V_j=\emptyset, 1\le i,j\le P$.} The end vertices of cross edges are called {\em boundary vertices}, and other vertices in $G$ are {\em inner vertices}. 
For a vertex-cut graph partition, the above $P$ partitions should satisfy $E=\cup E_i, E_i \cap E_j=\emptyset, 1\le i,j\le P$. A vertex may be replicated among partitions and the number of replications of vertex $v$ is called the replication factor. When the replication factor of a vertex is larger than 1, the vertex is called {\em  boundary vertex}, and other vertices in $G$ are {\em inner vertices}.

\textbf{Graph Neural Networks (GNNs)}. 
Given a graph $G$ with adjacent matrix $\bm{A}$ and the feature matrix $\bm{X}$ where each row is the initial feature vector $\bm{x_v}$ of a vertex $v$ in the graph $G$, an $l$-th layer in a GNN updates the vertex feature by aggregating the features from the corresponding neighborhoods, and the process can be formalized in matrix view as below
\begin{equation}
\bm{H^l} = \sigma(\tilde{\bm{A}}\bm{H^{l-1}W^{l-1}}),
\label{eq:gnnmatrix}
\end{equation}
where $\bm{H^l}$ is the hidden embedding matrix and $\bm{H^0}=\bm{X}$, $\bm{W^{l-1}}$ is the model weights, $\tilde{\bm{A}}$ is a normalized $\bm{A}$ and $\sigma$ is a non-linear function, e.g., Relu, Sigmoid, etc. 
Eq.~\ref{eq:gnnmatrix} is the global view of GNN computation. 
The local view of GNN computation is the computation of a single vertex. Given a vertex $v$, the local computation of $l$-th layer
in the message passing schema~\cite{gnnmp_icml_2017} can be formalized as below
\begin{equation}
\bm{h_v^{l}} = \sigma(\bm{h_v^{l-1}}, \underset{u\in N(v)}{\oplus}(\phi(\bm{h_v^{l-1}}, \bm{h_u^{l-1}}, \bm{h^{l-1}_{e_{u,v}}}))),
\label{eq:gnnmpi}
\end{equation}
where $\oplus$ is an aggregation function, $\phi$ is an update function, and $\bm{h^{l-1}_{e_{u,v}}}$ is the hidden embedding of edge $(u,v)$ at the $(l-1)$-th layer. 

\blue{
In contrast to traditional neural networks, where the data samples are typically independent and identically distributed (i.i.d.), GNN introduces data dependencies due to the underlying graph structure. These dependencies arise from the fact that the hidden embedding of each vertex in a GNN is updated based on the embeddings of its neighbors at the previous layer, which is formalized by Eq.~\ref{eq:gnnmatrix}  in global view and Eq.~\ref{eq:gnnmpi}  in local view, respectively. 
This dependency enables GNNs to capture the structural information in the graph. By leveraging the graph structure and capturing the dependencies between vertices and edges, GNNs are able to model complex relational data and perform tasks such as node classification and link prediction.
}

\blue{
\textbf{GNN Training Methods}. To train a GNN model, a basic solution is to update the model parameters once in a single epoch, such training method is called {\em full-graph GNN training}. However, full-graph GNN training is memory-intensive~\cite{gcn} since it requires to access the whole training graph and cannot scale to large graphs. 
Alternatively, we may choose to update the GNN model parameters multiple times over the course of a single epoch. This is known as {\em mini-batch GNN training}. An epoch is divided into multiple iterations and each iteration updates the GNN model with a batch (a.k.a, a subset of the whole training dataset).}

\begin{figure} 
    \centering
    \scalebox{0.5}{
        \includegraphics{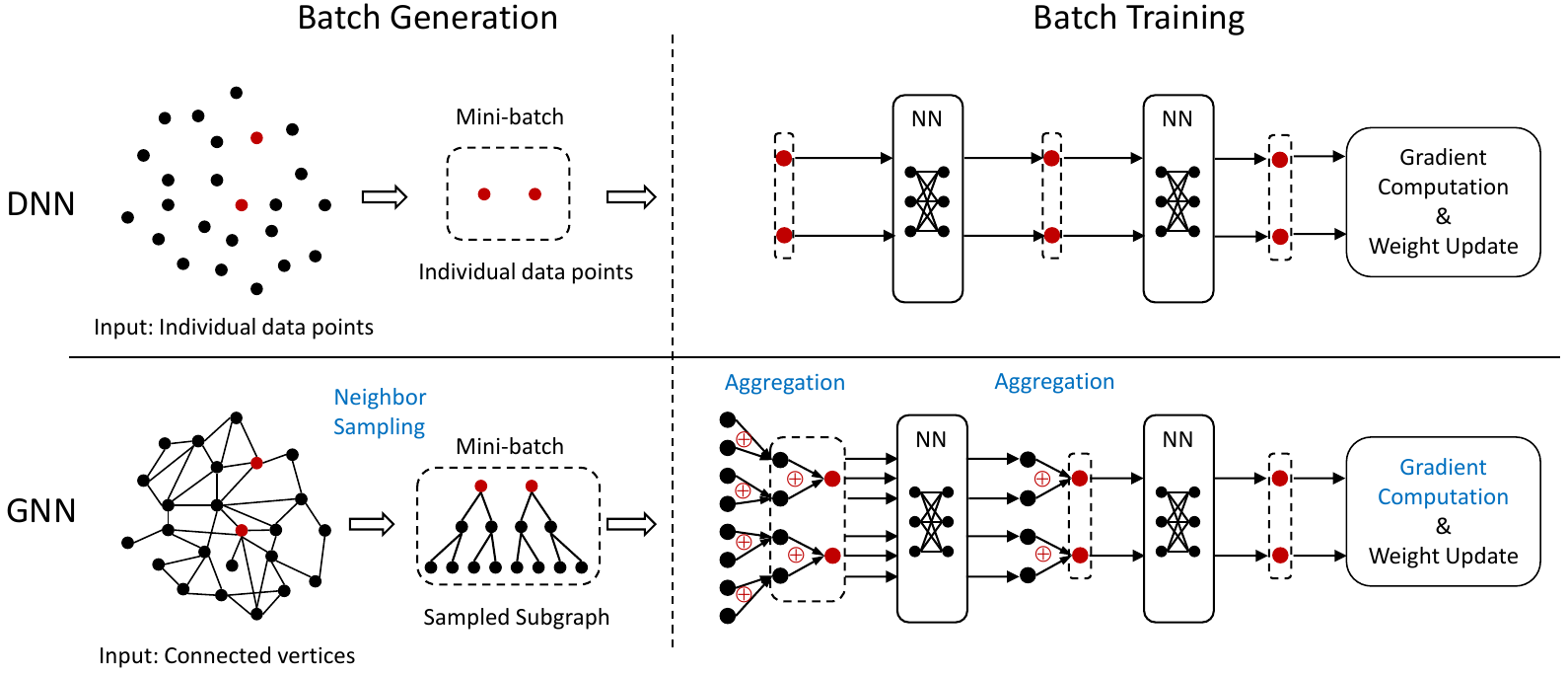}
    }
    \caption{\blue{An example of GNN execution. The difference between GNN and DNN is highlighted in \lhzblue{blue} color.}}
    \label{fig:gnn-example}
\vspace{-20pt}
\end{figure}

\blue{
Compared to traditional deep neural network (DNN) training, GNN training has two essential differences, which are illustrated by an example in Figure~\ref{fig:gnn-example}.
{First, for the batch generation in mini-batch training, GNN constructs a batch with $L$-hop neighbors involved, while traditional DNN involves the selected training samples only}. Because of the data dependency in GNN training, to compute the output embedding in the final layer (layer $L$) of a training vertex, we need all the embeddings of its neighbors from the previous layer (layer $L - 1$). Recursively, in order to compute the embeddings of neighboring vertices at layer $L - 1$, embeddings of its 2-hop neighbors are required. Therefore, an exact mini-batch containing only one training vertex involves all its $L$-hop neighbors, along with their features as the input of mini-batch training. 
A significant challenge arises when constructing such mini-batches: they can quickly become large and expand to include a substantial portion of the complete graph, even only a few vertices are selected as the central training vertices. Some methods are proposed to mitigate this problem by sampling in the neighborhood so to reduce the total size of the mini-batch, such as node-wise sampling~\cite{graphsage_nips_2017, vrgcn_icml_2018}, layer-wise sampling~\cite{fastgcn_iclr_2018, adaptive_nips_2018, ladies_nips_2019} and subgraph sampling~\cite{mvs_kdd_2020, graphsaint_iclr_2020}.
}

\blue{Second, for the model computation in both mini-batch training and full-graph training, an additional aggregation process is introduced, to collect vertex information from the neighborhood. After the output embeddings of training vertices are computed, a backward propagation is performed to compute the gradient for each parameter. Similar to the forward computation, the gradients on each vertex is sent alongside the edges to its neighbors during backward propagation, incurring an additional scatter process compared to DNN training. The aggregation and scatter operation among data samples leads to a much more complex computation process compared to DNN training.}

\blue{
\textbf{Distributed DNN Training}. Distributed DNN training~\cite{survey_distdnn_tpds20, survey_distdnn_ipdps21} is a solution to large-scale DNN. Parallelism and synchronization are two key components.
}
\blue{
\textbf{1) \textit{ Parallelism.}} Data parallelism is a prevalent training paradigm in distributed DNN training. In data parallelism, the model is replicated across multiple devices or machines, with each replica processing a different subset of the training data. Computed gradients are exchanged and averaged to synchronize the model parameters, often utilizing communication operations like all-reduce~\cite{allreduce_arxiv16}. To handle large models that exceed the capacity of a single GPU, model parallelism is adopted, where each device processes a distinct part of the model during forward and backward propagation.
}
\blue{
\textbf{2) \textit{Synchronization.}} Distributed training can be categorized into synchronous and asynchronous training. In synchronous training, all workers complete a forward and backward pass before model updates occur. This ensures that all workers are using the same model parameters for computation. In asynchronous training, workers update the model parameters independently, so to remove the global synchronization point between each mini-batch. Based on the idea of asynchronous training, pipeline parallelism is proposed~\cite{gpipe_nips19, pipedream_SOSP_2019} to perform flexible mini-batch training and improve resource utilization to accelerate the overall training time. 
}

\blue{Because of the large models in the DNNs, most efforts are made to manage the storage and synchronization of model parameters for the distributed DNN training. Since GNN models typically have shallow network structures, the management of model parameters is trivial. However, the input graphs of GNNs can be extremely large. The specific data dependency introduced by graph data structure significantly impacts the computation process of distributed GNN training, resulting in a substantial communication overhead that differs from DNN training, which becomes the main consideration and introduces new challenges (see Section~\ref{sec:challenges}). 
}

\section{Distributed GNN Training and Challenges} \label{sec:pipeline-challenges}


\subsection{General Distributed GNN Training Pipeline}
\label{sec:abstraction}
\begin{figure}[!h]
\centering
\resizebox{0.9\textwidth}{!}{
\includegraphics{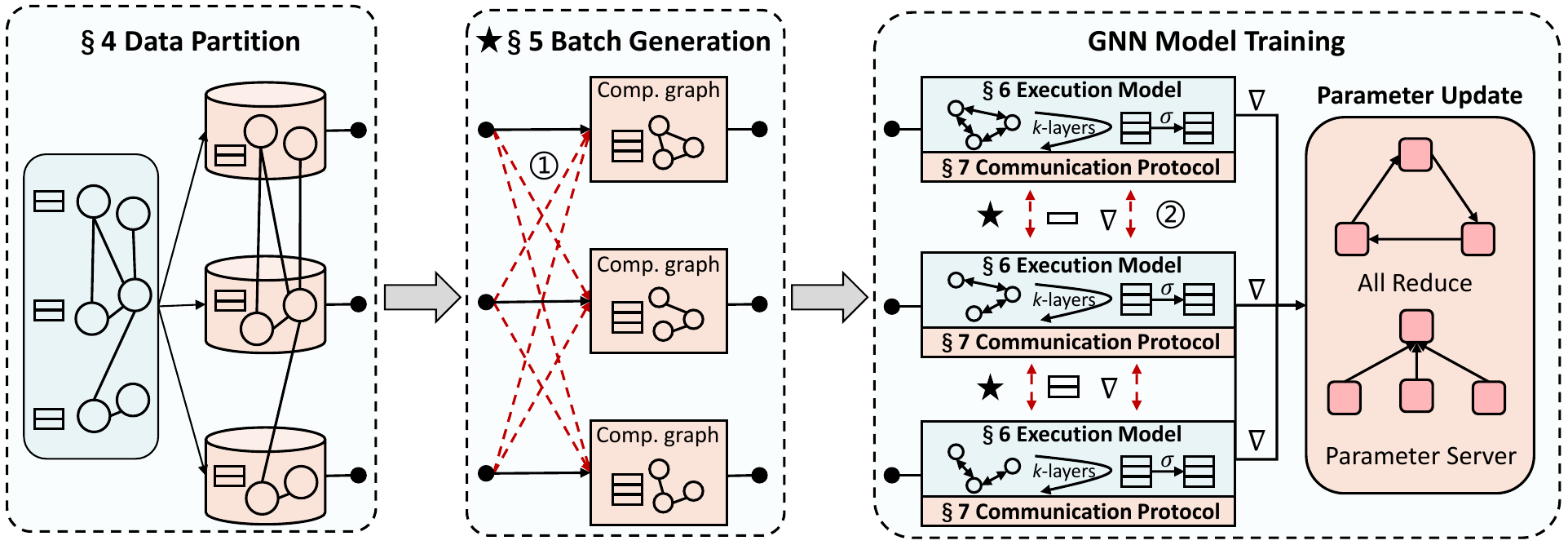}}
\vspace{-10pt}
\caption{\blue{The abstraction of distributed GNN training pipeline. Key stages are noted with the section number where they will be further discussed. Stages marked with star $\bigstar$ are optional for different training methods: batch generation is only involved in mini-batch training, and the communication protocol which is used to communicate hidden embeddings and gradients is involved in full-graph training.}}
\label{fig:gen_frame}
\vspace{-10pt}
\end{figure}

\blue{
To better understand the general workflow of end-to-end distributed GNN training, we divide the training pipeline into three stages that are data partition, batch generation and GNN model training. The GNN model training stage is further divided into model execution (including the design of the execution model and communication protocol) and parameter update. 
Figure~\ref{fig:gen_frame} visualizes the high-level abstraction of end-to-end distributed GNN training workflow. 
}

\blue{
\textbf{Data partition}. This is a preprocessing stage enabling distributed training. It distributes the input data (i.e., graph and feature) among a set of workers. Considering that the training data in GNN are dependent, the data partition stage becomes more complex than the one in traditional distributed machine learning. As shown in Figure~\ref{fig:gen_frame}, the cross-worker edges among the partitioned data (i.e., subgraphs) imply the data dependency. If we acknowledge the data dependency among partitions, the distributed training efficiency is reduced by communication; if we simply ignore the data dependency, the model accuracy is destroyed. Therefore, data partition is a critical stage for the efficiency of end-to-end distributed GNN training.
}


\blue{
\textbf{GNN batch generation.} This stage is only involved in distributed mini-batch training. In this stage, each worker generates a computation graph from the partitioned input graph and feature. However, due to the data dependency, the computation graph generation are quite different from the traditional deep learning model. The computation graph for the mini-batch training strategy may not be generated correctly without accessing remote input data. The remote data accessing makes the distributed batch generation more complicated than the traditional deep learning model.}

\blue{
\textbf{GNN model training.} This is the core stage for the embedding computing and model updating. It is further divided into model execution stage and parameter update stage. 
The model execution stage is specifically designed for distributed GNN training, while the parameter update stage is consistent to that of traditional DNN model training. We further divide the model execution stage into execution model and communication protocol. 
For distributed mini-batch training, only execution model is involved, which manages the scheduling of different training operators and how to overlap them efficiently with the batch generation stage. 
For distributed full-graph training, both execution model and communication protocol is included. The execution model involves $k$-layer graph aggregation of GNN models and the aggregation exhibits an irregular data access pattern. 
Furthermore, the graph aggregation in each layer needs to access the hidden embeddings and computed gradients of the remote neighbors via the communication protocol, and the synchronization schema between layers should be also considered. 
Finally, for parameter update stage, the existing techniques in classical distributed machine learning can be directly applied to the distributed GNN training.
In conclusion, the distributed GNN model training stage is more complicated than training traditional DNN training, and needs careful design for both the execution model and the communication protocol. 
}

\subsection{Challenges in Distributed GNN Training}
\label{sec:challenges}
Due to the data dependency, it is non-trivial to train distributed GNNs efficiently. We summarize three major challenges as below.

\textbf{Challenge \#1: Massive feature communication}. 
\blue{
For the distributed GNN training, massive feature communication is incurred in both the batch generation and GNN model training stage. 
When the GNN model is trained in a mini-batch manner, the batch generation needs to access remote graph and features as the {operation \textcircled{1} in Figure~\ref{fig:gen_frame}} to create a batch for local training.
}
Even if in multi-GPU training with a centralized graph store (e.g., CPU memory), the computation graph generation incurs massive data movement from the graph store to workers (e.g., GPU memory). Existing empirical studies demonstrate that the cost of mini-batch construction becomes the bottleneck during the end-to-end training~\cite{bytegnn_vldb_2022, nextdoor_eurosys_2021}. 
When the GNN model is trained in a full-graph manner, the computation graph can be the same as the partitioned graph without communication. However, the graph aggregation in each layer needs to access the hidden features in the remote neighbors of vertices (operation \textcircled{2} in Figure~\ref{fig:gen_frame}) which causes massive hidden feature (or embedding) communication. In summary, both distributed mini-batch training and distributed full-graph training of GNN models suffer from massive feature communication. 



\textbf{Challenge \#2: The loss of model accuracy}.
Mini-batch GNN training is more scalable than full-graph training, so it is adopted by most existing distributed GNN training  systems. However, with the increase of model depth, the exact mini-batch training suffers from the neighbor-explosion problem. A de-facto solution is to construct an approximated mini-batch by sampling or ignoring cross-worker edges. Although the approximated mini-batches improve the training efficiency, it cannot guarantee the model convergence in theory~\cite{roc_mlsys_2020}. Therefore, for the distributed mini-batch GNN training, we need to make a trade-off between model accuracy and training efficiency. In addition, there is a trend to develop full-graph distributed GNN training which has a convergence guarantee and is able to achieve the same model accuracy {compared to the local training on a single worker}.

\textbf{Challenge \#3: Workload imbalance}.
Workload balance is an intrinsic problem in distributed computing. However, the various workload characteristics of GNN models increase the difficulty to partition the training workload balance among workers. Because it is hard to model the GNN workload in a simple and unified way. Without formal cost models, the classical graph partitioning algorithms cannot be used to balance the GNN workload among workers.
In addition, distributed mini-batch GNN training requires that each worker handles the same number of mini-batches with the same batch size (i.e., subgraph size), not simply balancing the number of vertices in subgraphs.  In summary, it is easy to encounter workload imbalance when training GNN in a distributed environment, thus incurring workers waiting for each other and destroying the training efficiency. 

\begin{figure*}[!h]
\centering
\resizebox{0.9\textwidth}{!}{
\includegraphics{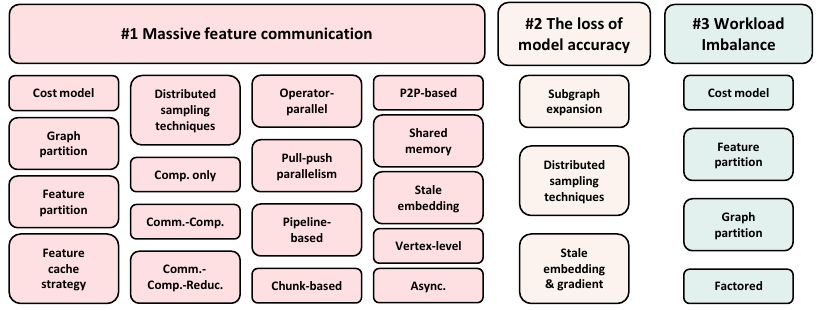}}
\vspace{-10pt}\caption{\blue{Challenges and techniques of distributed GNN training.}}
\label{fig:challenge}
\vspace{-10pt}
\end{figure*}

\subsection{Taxonomy of Distributed GNN Training Techniques} \label{sec:taxonomy}
\begin{figure*}[!h]
\centering
\resizebox{0.8\textwidth}{!}{
\includegraphics{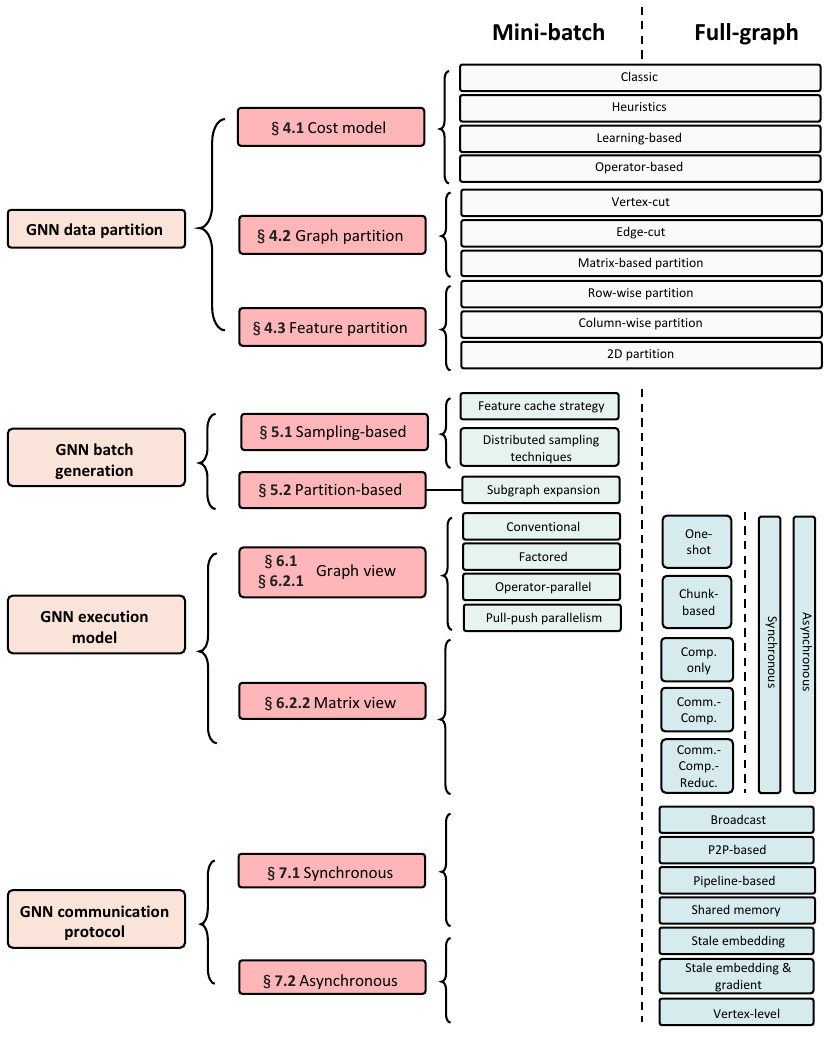}}
\vspace{-10pt}\caption{\blue{Taxonomy of distributed GNN training techniques}}
\label{fig:taxonomy}
\vspace{-10pt}
\end{figure*}

To realize the distributed GNN training and optimize the efficiency via solving the above challenges, many new techniques are proposed in the past years. Figure~\ref{fig:challenge} illustrates the relationship between challenges and the proposed techniques.
\blue{Previous works typically present their own contributions as a component of their proposed systems or frameworks, and may lack a comprehensive discussion of the techniques used in the same stage of distributed GNN training, or the techniques addressing the same challenge.}
In this survey, we introduce a new taxonomy by organizing the techniques specifically designed for distributed GNN training
on the basis of the stages in the end-to-end distributed training pipeline. With such a design, we organize similar techniques that optimize the same stage in the distributed GNN training pipeline together and help readers fully understand the existing solutions for the different stages in distributed GNN training.

{According to the previous empirical studies, due to the data dependency, the bottleneck of distributed GNN training generally comes up in the stages of data partition, batch generation and GNN {model training}, as shown in the pipeline. Furthermore, various training strategies (e.g., mini-batch training, full-graph training) bring in different workload patterns and result in different optimization techniques used in batch generation stage and {model training} stage. For example, the computation graph generation in batch generation stage is important to mini-batch training while communication protocol is important to full-graph training. In consequence, our new taxonomy classifies \blue{the techniques specifically designed for distributed GNN training}
into four categories (i.e., GNN data partition, GNN batch generation, GNN execution model and GNN communication protocol) as shown in Figure~\ref{fig:taxonomy}. In the following, we introduce the overview of each category.}
\textbf{GNN data partition}. In this category, we review the data partition techniques for distributed GNN training. The goal of data partition is to balance the workload and minimize the communication cost for a GNN workload. GNN training is an instance of distributed graph computing, many traditional graph partition methods can be directly used. However, they are not optimal for distributed GNN training because of the new characteristics of GNN workloads. Researchers have paid much effort to design GNN-friendly cost models which guide traditional graph partition methods. In addition, the graph and feature are two typical types of data in GNN and both of them are partitioned. Some works decouple the features from the graph structure and partition them independently. {In Section~\ref{sec:data-partition}, we elaborate on the existing GNN data partition techniques.}




\textbf{GNN batch generation}. In this category, we review the techniques of GNN batch generation for mini-batch distributed GNN training. The methods of mini-batch generation affect both the training efficiency and model accuracy. Graph sampling is a popular approach to generate a mini-batch for large-scale GNN training. However, the standard graph sampling techniques do not consider the factors of distributed environments and each sampler on a worker will frequently access data from other workers incurring massive communication. Recently, several new GNN batch generation methods optimized for distributed settings have been introduced. We further classify them into {distributed sampling mini-batch generation and partition-based mini-batch generation}. In addition, the cache has been extensively studied to reduce communication during the GNN batch generation. {In Section~\ref{sec:batch-generation}, we elaborate on the existing GNN batch generation techniques.}

\textbf{GNN execution model}. 
\blue{In this category, we review the execution model for both 
mini-batch and full-graph GNN training.
In mini-batch GNN training, 
sampling and feature extraction are two main operations that dominate the total training time. To improve efficiency, different execution models are proposed to schedule the training stage of a mini-batch with the sampling and feature extraction stage, in order to fully utilize computing resources. 
In full-graph GNN training, 
the neighbors' state for each vertex is aggregated in the forward computation and the gradients are scattered back to the neighbors in the backward computation, leading to massive communication of remote vertices. 
Due to the data dependency and irregular computation pattern, traditional machine learning parallel models (e.g., data parallel, model parallel, etc.) are not optimal for graph aggregation, especially when feature vectors are high-dimensional. 
We introduce the full-graph execution model from both matrix view and graph view. From matrix view, we classify the execution of distributed SpMM into computation-only, communication-computation, and communication-computation-reduction models. From graph view, we classify the execution models from two perspectives. From the perspective of the execution orders of graph operators, we classify the full-graph execution model into one-shot execution model and chunk-based execution model.
From the perspective of the synchronization timeliness in graph aggregation, we classify the full-graph execution models into synchronous execution model and asynchronous execution model. Based on the removal of different synchronization points, the asynchronous execution model can be further categorized into two types. A detailed description of these techniques is presented in Section~\ref{sec:exec-model}.
}



\textbf{GNN communication protocol}. 
In this category, we review the communication protocols of distributed full-graph parallel training. The graph aggregation of distributed full-graph training needs to access remote hidden embedding access because the neighbors of a vertex cannot be always at local with graph partitioning. Based on the synchronous and asynchronous execution model in graph aggregation, these communication protocols can also be categorized as synchronous communication protocol and asynchronous communication protocol, respectively.
The former assumes in each layer, the model access the latest embedding while the latter assumes in each layer the model is allowed to access {stale} embedding (or historical embeddings). A detailed description of various communication protocols is presented in Section~\ref{sec:communication-protocol}.

\section{GNN Data Partition} \label{sec:data-partition}

\begin{figure}[!h]
\centering
\scalebox{0.4}{
\includegraphics{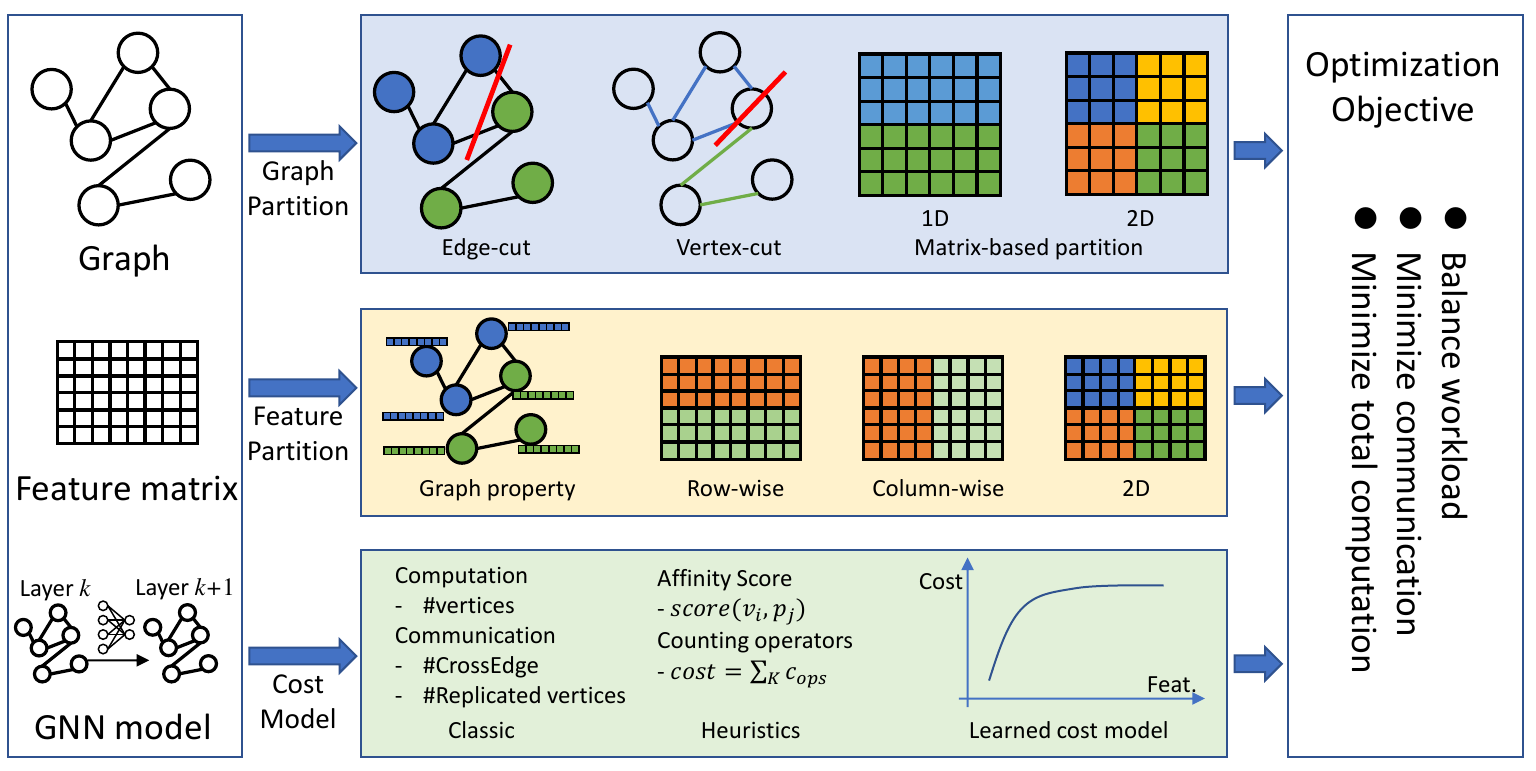}}
\caption{GNN data partition techniques}
\label{fig:parttech}
\vspace{-10pt}
\end{figure}


In this section, we review existing techniques of GNN data partition in distributed GNN training. Figure~\ref{fig:parttech} describes the overview of the techniques. Considering graphs and features are two typical types of data in GNN, we classify partition methods into graph partition and feature partition. The optimization objectives are workload balance and communication and computation minimization, which aim at addressing challenges \#1 and \#3. 
\blue{
In addition, the cost model is another critical component that captures the characteristics of GNN workloads. With a well-designed cost model, workloads on each partition can be accurately estimated. A graph partition strategy can better address the challenges in GNN by adopting a more accurate cost model.
In the following, we first present various cost models, which are the basis of graph partition. Then we discuss the graph partition and feature partition, respectively.
}

\subsection{Cost model of GNN}
\blue{
The cost model estimates the computation and communication cost of a GNN workload. Therefore, building a cost model is an important technique that underpins the data partition.
} 
In general, for a graph analysis task, we use the number of vertices to estimate the computation cost and the number of cross-edges to estimate the communication cost~\cite{gp_metis}. However, this simple method is not proper for the GNN tasks because they are not only influenced by the number of vertices and cross-edges, but also influenced by the dimension of features, the number of layers, and the distribution of training vertices. 
Researchers have proposed several GNN-specific cost models, including the heuristics model, learning-based model, and operator-based model.

\textbf{Heuristics model} selects several graph metrics to estimate the cost via simple user-defined functions. Several heuristics models for GNN workloads have been introduced in the context of streaming graph partition~\cite{stream_part_kdd, stream_part_tkde}, which assigns a vertex or block to a partition one by one. Such models define an affinity score for each vertex or block and the score helps the vertex or block select a proper partition. 

{Assume a GNN task is denoted by $GNN(L, G,$ $V_{train}$, $V_{valid}$, $V_{test})$, where $L$ is the number of layers, $G$ is the graph, $V_{train}$, $V_{valid}$ and $V_{test}$ are the vertex set for train, validation and test. 
The graph $G$ is partitioned into $K$ subgraphs. In the context of streaming graph partition, for each assignment of vertex or block, let $P_i$ $(1 \le i \le K)$ be the set of vertices that have been already assigned to it, and $V_{train}^{i}$, $V_{valid}^i$ and $V_{test}^i$ are the corresponding vertex set belong to partition $P_i$.
}



Lin et al.~\cite{pagraph_socc_2020} define an affinity score vector with $K$ dimension for each train vertex $v_t \in V_{train}$, in which each score represents the affinity of the vertex to a partition. Let $IN(v_t)$ be the $L$-hop in-neighbor set of train vertex $v_{t}$,  
the score of $v_t$ with respect to partition $P_i$ is defined as below,
\begin{equation}
score_{v_t}^{i}=|V_{train}^{i}\cap IN(v_{t})|\cdot\frac{V_{train}^{avg}-|V_{train}^{i}|}{|P_{i}|},
\label{eq:pagraph}
\end{equation}
where $V_{train}^{avg} = \frac{V_{train}}{K}$. This score function implicitly balances the number of training vertices among partitions. Liu et al.~\cite{bgl_nsdi23} define similar affinity score with respect to a block (or subgraph) $B$, and the formal definition is



\begin{equation}
|P_{i}\cap IN(B)|\cdot(1-\frac{|P_{i}|}{P_{avg}})\cdot(1-\frac{|V_{train}^{i}|}{V_{train}^{avg}}),
\label{eq:bgl}
\end{equation}
where $P_{avg} = \frac{|V|}{K}$.
Zheng et al.~\cite{bytegnn_vldb_2022} define the affinity score of a block $B$ by considering all the training, validation and test vertices, the formula is 
\begin{equation}
\frac{CrossEdge(P_{i}, B)}{|P_{i}|}\cdot(1-\alpha \frac{V_{train}^{i}}{V_{train}^{avg}} -\beta \frac{V_{valid}^{i}}{V_{valid}^{avg}} -\gamma \frac{V_{test}^{i}}{V_{test}^{avg}}),
\label{eq:bytegnn}
\end{equation}
where $CrossEdge(P_i, B))$ is the number of cross-edges between $B$ and $P_{i}$, $\alpha, \beta, \gamma$ are hyper-parameters manually set by users.


\textbf{Learning-based model} takes advantage of machine learning techniques to model the complex cost of GNN workloads. The basic idea is to manually extract features via feature engineering and apply classical machine learning to train the cost model. The learning-based model is able to estimate the cost using not only the static graph structural information but also runtime statistics of GNN workloads, thus achieving a more accurate estimation than the heuristic models. 
Jia et al.~\cite{roc_mlsys_2020} introduce a linear regression model for GNN computation cost estimation. 
The model estimates the computation cost of a single GNN layer $l$ with regard to any input graph $G=(V, E)$. For each vertex in the graph, they select five features (listed in Table~\ref{tbl:roc_features}), including three graph-structural features and two runtime features. The estimation model is formalized as below,
\begin{equation}
t(l,v)=\sum\limits_{i}w_{i}(l)x_{i}(v)
\end{equation}
\begin{equation}
\begin{aligned}
t(l,G)&=\sum\limits_{v \in V}t(l, v)=\sum\limits_{v \in V}\sum\limits_{i}w_{i}x_{i}(v)\\&=\sum\limits_{i}w_{i}\sum\limits_{v \in V}x_{i}(v)=\sum\limits_{i}w_{i}x_{i}(G)
\end{aligned}
\label{eq:roc}
\end{equation}
where $w_{i}(l)$ is a trainable parameter for layer \emph{l}, $x_{i}(v)$ is the \emph{i}-th feature of \emph{v}, and $x_{i}(G)$ sums up \emph{i}-th feature of all vertices in \emph{G}.

\begin{table}
\caption{The vertex features used in ROC's cost model}
\label{tbl:roc_features}
\resizebox{0.9\textwidth}{!}{
\begin{tabular}{|l|l|l|l|}
\hline
   Name & Type & Definition & Description \\ \hline
$x_{1}$ & graph-structural feature &1          & the vertex itself            \\ 
$x_{2}$ & graph-structural feature &$|N(v)|$           & the number of neighbors            \\ 
$x_{3}$ & graph-structural feature &$|C(v)|$           & continuity of neighbors            \\ 
$x_{4}$ & runtime feature &$\sum_{i} \lceil \frac{c_{i}v}{WS} \rceil$           & mem accesses to load neighbors            \\ 
$x_{5}$ & runtime feature & $\sum_{i} \lceil \frac{c_{i}v \times d_{in} }{WS}\rceil$           & mem accesses to load the activations of all neighbors            \\ \hline
\end{tabular}
}
\end{table}



Wang et al.~\cite{flexgraph_eurosys_2021} use a polynomial function $f$ to estimate the computation cost of vertex over a set of manually selected features. The formal definition is 
\begin{equation}
f=\sum\limits_{i=0}^{T}n_{i}m_{i},
\label{eq:flexgraph}
\end{equation}
where $T$ is the number of neighbor types (e.g., a metapath~\cite{flexgraph_eurosys_2021}) which is defined by the GNN models, $n_i$ is the number of neighbors for the $i$-th type, $m_i$ is the total feature dimensions of the $i$-th type of neighbor instance (i.e., a neighbor instance of the $i$-th type has $n$ vertices and each vertex has feature dimension $f$, then $m_i=n\times f$). {Refer to the original work~\cite{flexgraph_eurosys_2021} for the detailed examples of the function $f$.} The total computation cost of a subgraph is the sum of the estimated costs of the vertices in the subgraph.

\textbf{Operator-based model} enumerates the operators in a GNN workload and estimates the total computation cost by summing the cost of each operator. Zhao et al.~\cite{distgnn_CMGCN} divide the computation of a GNN workload into forward computation and backward computation. The forward computation of a GNN layer is divided into aggregation, linear transformation, and activation function; while the backward computation of a GNN layer is divided into gradient computation towards loss function, embedding gradient computation, and gradient multiplications. As a result, the costs of computing embedding $h_{v}^{l}$ of vertex $v$ in layer $l$ in forward and backward propagation are estimated by $c_{f}(v, l)$ and $c_b(v,l)$, respectively.
\begin{equation}
 c_{f}(v, l)=\alpha|N_{v}|d_{l-1}+\beta d_{l}d_{l-1}+\gamma d_{l},  
\end{equation}
\begin{equation}
c_b(v,l)=\left\{
\begin{array}{ll}
(\lambda + \eta)d_{l} + (2\beta + \eta)d_{l}d_{l-1}, &  l=L \\
\alpha|N_{v}|d_{l} + (\beta + \eta)d_{l}d_{l-1} + \eta d_{l}, &  0<l<L
\end{array}
\right.
\end{equation}
where $d_l$ is the dimension of hidden embeddings in the $l$-th GNN layer, $|N_v|$ is the number of neighbors of vertex $v$, $\alpha$, $\beta$, $\gamma$, and $\eta$ are constant factors which can be learned through testing the running time in practice. 
Finally, the computation cost of a mini-batch $B$ is computed by summing up the computation cost from all vertices in the $B$ from layer 1 to layer $L$ as below,
\begin{equation}
C(B)=\sum\limits_{l=0}^{L-1}\sum\limits_{v \in \bigcup_{u \in B}N_{u}^{l}}(c_{f}(v,l)+c_{b}(v,l)),
\label{eq:cmgcn}
\end{equation}
where $N_{u}^{l}$ represents the vertices in graph \emph{G} that are \emph{l}-hop away from vertex \emph{u}.


\subsection{Graph partition in GNN}
\label{sec:gp}
GNN is a type of graph computation task. Classical graph partition methods can be directly applied to the GNN workloads. 
Many distributed GNN training algorithms and systems like AliGraph~\cite{aligraph_vldb_2019}, DistGNN~\cite{distgnn_sc_2021}  adopt METIS~\cite{gp_metis}, Vertex-cut~\cite{powergraph}, Edge-cut~\cite{gp_metis}, streaming graph partition~\cite{stream_part_kdd}, and other graph partition methods~\cite{gp_matrix} to train the GNN models in different applications. Recently, Tripathy et al.~\cite{partition_sc_2020} empirically studied the performance of 1-D, 2-D, 1.5-D, and 3-D graph partitioning methods with respect to GCN models. However, due to the unique characteristics of GNN workloads, classical graph partition methods are not optimal to balance the GNN workload while minimizing the communication cost.

In distributed mini-batch GNN training, to balance the workload, we need to balance the number of train vertices (i.e., the number of batches per epoch) among workers. As a result, new optimization objectives are introduced.
Zheng et al.~\cite{distdgl_ai3_2020} formulate the graph partition problem in the DistDGL as a multi-constraint partition problem that aims at balancing training/validation/test vertices/edges in each partition. 
{They adopt the multi-constraint mechanism in METIS to realize the customized objective of graph partition for both homogeneous graphs and heterogeneous graphs~\cite{distdglv2_kdd_2022}, and further expand this mechanism to a two-level partitioning strategy that handles the situation with multi-GPUs and multi-machines and balances the computation.}
Lin et al.~\cite{pagraph_socc_2020} apply streaming graph partition -- Linear Deterministic Greedy (LGD)~\cite{stream_part_kdd} and use a new affinity score (E.q.~\ref{eq:pagraph}) to balance the GNN workload. Furthermore, they ensure each vertex in a partition has its complete $L$-hop neighbors to avoid massive communication during sampling. 
Recently, block-based streaming graph partition methods~\cite{bgl_nsdi23, bytegnn_vldb_2022} are introduced for mini-batch distributed GNN training. They first use multi-source BFS to divide the graph into many small blocks, then apply greedy assignment heuristics with customized affinity scores (E.q.~\ref{eq:bgl},~\ref{eq:bytegnn}) to partition the coarsened graph, and uncoarsen the block-based graph partition by mapping back the blocks to the vertices in the original graph. Besides workload balance, some works minimize the total computation cost. Zhao et al.~\cite{distgnn_CMGCN} prove that the computation cost function (E.q.~\ref{eq:cmgcn}) is a submodular function and use METIS to partition the graph into cohesive mini-batches achieving $2-\frac{2}{M}$ approximation of the optimal computation cost.


In distributed full-graph GNN training, we need to balance the workload of each GNN layer among a set of workers while minimizing the embedding communication among workers. 
\blue{
One approach is to leverage a heuristic method to model the computation workload and the communication overhead, and adopt the multi-constraint graph partitioning mechanism. 
Wan et al.~\cite{g3-sigmod23} models the computation workload as both the number of vertices and the number of edges, and treat them as two constraints. The communication overhead is modeled as the number of remote vertices in each worker. Therefore, with the multi-constraint graph partitioning, the computation workload can be balanced and the communication overhead can be reduced. To further balance the communication overhead, an iterative re-partitioning stage is adopted to swap vertices between workers with the highest and lowest communication volumes. Demirci et al.~\cite{spmm-vldb23} model the computation workload as the total number of vertices' neighbors in each worker. To model the communication overhead under P2P-based communication protocol (introduced in Section~\ref{sec:sync_comm_protocol}), they introduce the hypergraph to model the communication cost of each vertex as the connectivity of its corresponding net in the hypergraph. 
Based on the hypergraph model, existing methods~\cite{hypergraph_partition_tpds99} for hypergraph partitioning is adopted to distribute vertices on different workers. 
}
\blue{Another approach is to apply learning-based cost models to model the complex cost for diverse GNN models.}
On top of the model-aware cost models, existing graph partition methods are adopted. For example, 
Jia et al.~\cite{roc_mlsys_2020} utilize the linear-regression-based cost model (E.q.~\ref{eq:roc}) and apply range graph partitioning to balance the workload estimated by the cost model. In addition, the range graph partition guarantees that each partition holds consecutively numbered vertices, thus reducing the data movement cost between CPU and GPU. 
Wang et al.~\cite{flexgraph_eurosys_2021} use the E.q.~\ref{eq:flexgraph} to estimate the computation cost of a partition, and apply the application-driven graph partition method~\cite{adp_sigmod20} to generate workload balancing plan, which adaptively mitigates workload and minimizes the communication cost. 
Based on the vertex-cut partition, Hoang et al.~\cite{deepgalois_mlsys21} leverage the 2D Cartesian vertex cut to improve the scalability. 



\subsection{Feature partition in GNN}
As an important type of data in GNN, features need to be partitioned as well. Most distributed GNN training solutions treat features as the vertex properties of a graph and they are partitioned along with the graph partition. For example, if the graph is partitioned by the edge-cut method, then each vertex feature is stored in the partition where the corresponding vertex lies at. In other words, the feature matrix is partitioned in row-wise. 

Considering the different processing patterns of features compared to the graph structure, a line of works partition feature matrix independent of the graph. When a graph is partitioned by the 2D partition method, Tripathy et al.~\cite{partition_sc_2020} partition the feature matrix with the 2D partition method as well, while Vasimuddin et al.~\cite{distgnn_sc_2021} use row-wise method to partition the feature matrix with replication and ensure each vertex has the complete feature at local.
Gandhi et al.\cite{p3_osdi_21} deeply analyze the communication pattern in distributed mini-batch GNN training and find that classical graph partition methods only reduce the communication cost at the first hop process. Furthermore, sophisticated graph partitioning, e.g., METIS, is an expensive preprocess step. They introduce to partition the graph and feature separately. The graph is partitioned via random partition to avoid the expensive preprocess. The input features are partitioned along the feature dimension (a.k.a. column-wise partition). In other words, each partition contains a sub-column of all vertices' features. 
{With such data partition strategy, they further design a new execution model introduced in Section~\ref{sec:async-data-parallel} and train the GNN model with minimal communication as described in Section~\ref{sec:asyn-protocol-realization}, especially, when the input feature has high dimension and the hidden features have low dimension.}
Similarly, Wolf et al.~\cite{distgnn_GIST} also use a column-wise feature partition method to distributed training for ultra-wide GCN models. Differently, in this method, the graph is partitioned with METIS and both input and intermediate features are partitioned along the feature dimension.

\section{GNN Batch Generation}\label{sec:batch-generation}
\begin{figure}[!h]
\centering
\scalebox{0.55}{
\includegraphics{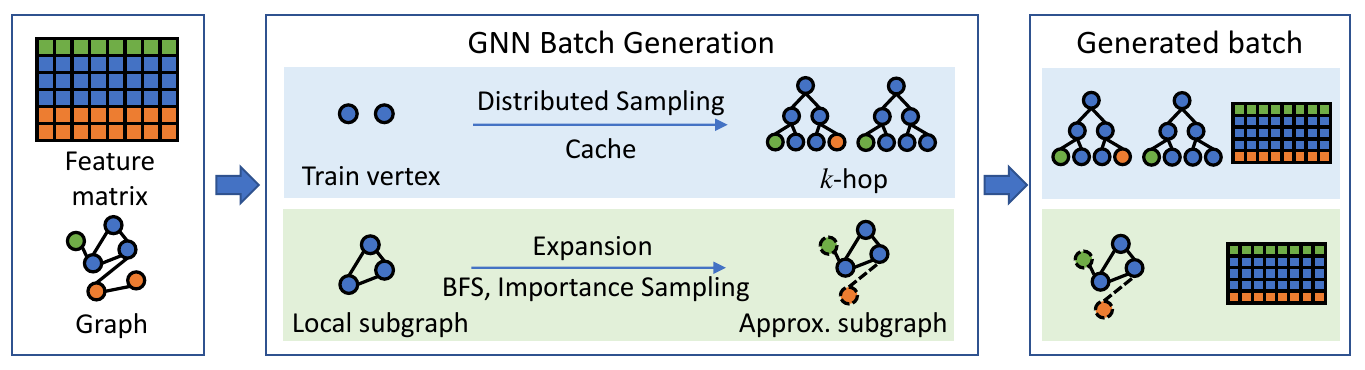}}
\vspace{-10pt}
\caption{GNN batch generation techniques}
\label{fig:batchgen}
\vspace{-10pt}
\end{figure}
Mini-batch GNN training is a common approach to scaling GNNs to large graphs.
Graph sampling is a de-facto tool to generate mini-batches in standalone mode. So far, many sampling-based GNNs~\cite{graphsage_nips_2017, fastgcn_iclr_2018, adaptive_nips_2018, ladies_nips_2019, graphsaint_iclr_2020, mvs_kdd_2020} have been proposed, and they can be classified into vertex-wise sampling, layer-wise sampling and subgraph-wise sampling according to different types of sampling methods. 
Different batch generation methods influence both the training efficiency and training accuracy. To avoid the graph sampling becoming a bottleneck, there are some explorations for the efficient GNN dataloaders~\cite{gns_kdd_2021,2021_VLDB_GPU-oriented-datacomm,2021_TPDS_Efficient_dataloader,2019_arxiv_tigergraph}.

In mini-batch distributed GNN training, the data dependence brings in massive communication for the batch generation process. To improve the training efficiency in distributed settings, several new GNN batch generation techniques that are specific to distributed training are proposed and address challenge \#1 and challenge \#2. As shown in Figure~\ref{fig:batchgen},
one solution is generating a mini-batch via distributed sampling and the other is directly using the local partition (or subgraph) as the mini-batch.
In the following, we describe these techniques.

\subsection{Mini-batch generation with distributed sampling}
Based on existing sampling-based GNNs, it is straightforward to obtain the distributed version by implementing distributed sampling. In other words, we use distributed graph sampling to create mini-batches over large graphs.
Most of existing distributed GNN systems, like AliGraph~\cite{aligraph_vldb_2019}, DistDGL~\cite{distdgl_ai3_2020}, BGL~\cite{bgl_nsdi23}, follow this idea. However, in a distributed environment, a basic sampler on a worker will frequently access data from other workers incurring massive communication. In multi-GPU settings, although graphs and features might be centralized and stored in CPU memory, each GPU needs to access the mini-batch from CPU memory via massive CPU-GPU data movement.

To reduce communication in distributed computing, cache is a common optimization mechanism for improving data locality. At the system level, many GNN-oriented cache strategies have been proposed. 
The basic idea of caching is to store frequently accessed remote vertices at local. 
Zheng et al.~\cite{distdgl_ai3_2020} propose to replicate the remote neighbors of boundary vertices at local for each partitioned graph in DistDGL and ensure each vertex at local has the full neighbors so that the mini-batch generator (e.g., sampler) can compute locally without communicating to other partitions. However, such a method only reduces the communication caused by the direct (i.e., one-hop) neighbor access and is unaware of the access frequency. Zhu et al.~\cite{aligraph_vldb_2019} introduce a metric called $l$-th importance of $v$, denoted by $Imp^{(l)}(v)$. The metric is formally defined as 
$Imp^{(l)}(v)=\frac{D_{i}^{(l)}(v)}{D_{o}^{(l)}(v)}$,
where $D_{i}^{(l)}(v)$ and $D_{o}^{(l)}(v)$ are the number of $l$-hop in and out-neighbors of the vertex $v$.
To balance the communication reduction and storage overhead, they only cache the vertices when its $Imp^{(l)}(v)$ is larger than a threshold. 
Li et al.~\cite{pagraph_socc_2020} design a static GPU cache that stores the high out-degree vertices for the multi-GPU training. The static cache can reduce the feature movement between CPU and GPU. Min et al.~\cite{datatiering} introduce weighted reverse PageRank (Weighted R-Pagerank)~\cite{rpagerank} that incorporates the labeling status of the vertices into the reverse PageRank method to identify the frequent access vertices. The hot vertex features with a high score are cached in GPU memory and the rest of the cold vertex features are placed in CPU memory. Furthermore, due to the different bandwidths between NVLink and PCIe, they divided the hot vertex features into the hottest one and the next hottest one. The hottest vertex features are replicated over multiple GPUs. The next hottest data is scattered over multiple GPUs and they can be accessed from the peer GPU memory.
\blue{
While the above strategies only focus on feature cache, latest studies~\cite{ducati_sigmod23, legion_atc23} find that when GPU sampling is used, caching both the vertex features and graph topology achieves better performance. To find the optimal cache allocation for graph topology and vertex features, Sun et al.~\cite{legion_atc23} leverage a linear search strategy to obtain the ratio of the two cache sizes, and allocate each cache with the top hottest data. In contrast, Zhang et al.~\cite{ducati_sigmod23} model this allocation problem as a knapsack problem, and design a greedy algorithm to find the optimal cache allocation strategy.  
}

The above static cache strategies cannot be adaptive to various graph sampling workloads.  
Liu et al.~\cite{bgl_nsdi23} propose a dynamic cache mechanism that applies the FIFO policy by leveraging the distribution of vertex features. To improve the cache hit ratio, they introduce proximity-aware ordering for a mini-batch generation. The order constrains the training vertex sequence in BFS order. To ensure model convergence, they introduce randomness to the BFS-based sequence by selecting a BFS sequence in a round-robin fashion and randomly shifting the BFS sequence. 
Yang et al.~\cite{gnnlab} propose a more robust cache approach, called a pre-sampling-based caching policy. 
Given a graph \emph{G}, a sampling algorithm \emph{A}, and a training set \emph{T},
it first executes $K$ sampling stages to collect statistics, then uses the statistics to identify the frequency (or hotness) of the data, finally caches the data with larger hotness. 
\blue{
Kayer et al.~\cite{salient++-mlsys23} adopt the same idea and propose an analysis-based cache policy that performs more accurate prediction on the data hotness. A propagation model is designed in the analysis method, and it iteratively propagates and calculates the sampled probability of each vertex given \emph{G}, \emph{A}, and \emph{T}. 
Both the pre-sampling-based and analysis-based caching policies consider all three factors that influence the data access frequency, and they are more robust than previous cache strategies for the various GNN workloads.
}

Besides the cache strategy, another approach is to design new distributed sampling techniques that are communication-efficient and maintain the model accuracy. A basic idea of communication-efficient sampling is to {prioritize sampling} local vertex, {yet this} introduces bias for the generated mini-batch. 
Inspired by the linear weighted sampling methods~\cite{fastgcn_iclr_2018, ladies_nips_2019}, 
Jiang et al.~\cite{distgnn_commefficient} propose a skewed linear weighted sampling for the neighbor selection. 
Concretely, the skewed sampling scales the sampling weights of local vertices by a factor $s > 1$ and Jiang et al. theoretically prove that the training can achieve the same convergence rate as the linear weighted sampling by properly selecting the value of $s$. 
\blue{
Cai et al. propose a technique called collective sampling primitive (CSP)~\cite{dsp-ppopp23}. CSP reduces communication costs by pushing the sampling tasks to remote workers instead of pulling the entire neighbor lists to the local worker. This is beneficial because the full neighbor list of a center vertex is often much larger than the sampled results. With CSP, sampling tasks for remote vertices are conducted on remote workers, and only the results are sent back, reducing communication overhead.
}

\blue{
In addition to the above GNN-specific distributed sampling techniques, other general distributed graph sampling methods on CPU clusters or GPU clusters can also provide help, such as C-SAW~\cite{csaw_sc_2020}, Knightking~\cite{knightking_sosp19} and Skywalker~\cite{skywalker-pact21}.
}

\subsection{Partition-based mini-batch generation}
According to the abstraction of distributed GNN training pipeline in Figure~\ref{fig:gen_frame}, we know that the graph is partitioned at first. If we restrict each worker to only training the GNN model with the local partition, massive communication can be avoided. In such a training strategy, a partition is a mini-batch, and we call it a partition-based mini-batch. 
PSGD-PA~\cite{distgnn_llcg} is a straightforward implementation of the above idea with a Parameter Server. In GraphTheta~\cite{graphtheta}, the partitions are obtained via a community detection algorithm.

However, the ignorance of cross edges among partitions brings in the loss of model accuracy. 
To improve the model accuracy, subgraph (i.e., partition) expansion is introduced. It preserves the local structural information of boundary vertices by replicating remote vertices to local. 
Xue et al.~\cite{sugar} use METIS to obtain a set of unoverlapped subgraphs, and then expand each subgraph by adding the one-hop neighbors of vertices that do not belong to the subgraph. Similarly, Angerd et al.~\cite{distsub} use the breadth-first method to expand the $l$-hop subgraph. Zhao et al.~\cite{gad} define a Monte-Carlo-based vertex importance measurement and use a depth-first sampling to expand the subgraph. They further introduce variance-based subgraph importance and weighted global consensus to reduce the impact of subgraph variance and improve the model accuracy. In addition, Ramezani et al.~\cite{distgnn_llcg} introduced \blue{LLCG (Learn Locally, Correct Globally)} to improve the model accuracy for the partition-based mini-batch training. LLCG follows the Parameter Server architecture and periodically executes a global correction operation on a parameter server to update the global model. The empirical results demonstrate that the global correction operation can improve the model accuracy significantly.

\section{GNN Execution Model} \label{sec:exec-model}
\blue{The execution model is the core stage for GNN model training, as shown in Figure~\ref{fig:gen_frame}. It is responsible to schedule different operations to achieve high training efficiency. 
For distributed mini-batch training, as the sampling and feature extraction operations in batch generation dominate the training efficiency and make the computation graph generation expensive, the execution model focuses on how to efficiently schedule these operations with the batch training operation. 
While for distributed full-graph training, the computation graph execution is complicated due to the data dependency among workers, and the execution model focuses on the scheduling of graph aggregation and communication during computation graph execution. 
}

\blue{In the following, due to different execution manners of mini-batch and full-graph GNN training, we discuss the execution model of them respectively.}

\subsection{Mini-batch Execution Model}
\label{exec_generation}
\blue{
In distributed mini-batch GNN execution, sampling and feature extraction are two unique operations which become the bottleneck. {As shown in \cite{bgl_nsdi23, p3_osdi_21}, the sampling and feature extraction takes up almost 83\% - 99\% of the overall training time.} The low efficiency is caused by massive communication and resource contention. Therefore, several execution models for mini-batch execution are proposed.}

\begin{figure}
    \centering
    \resizebox{0.8\textwidth}{!}{
        \includegraphics{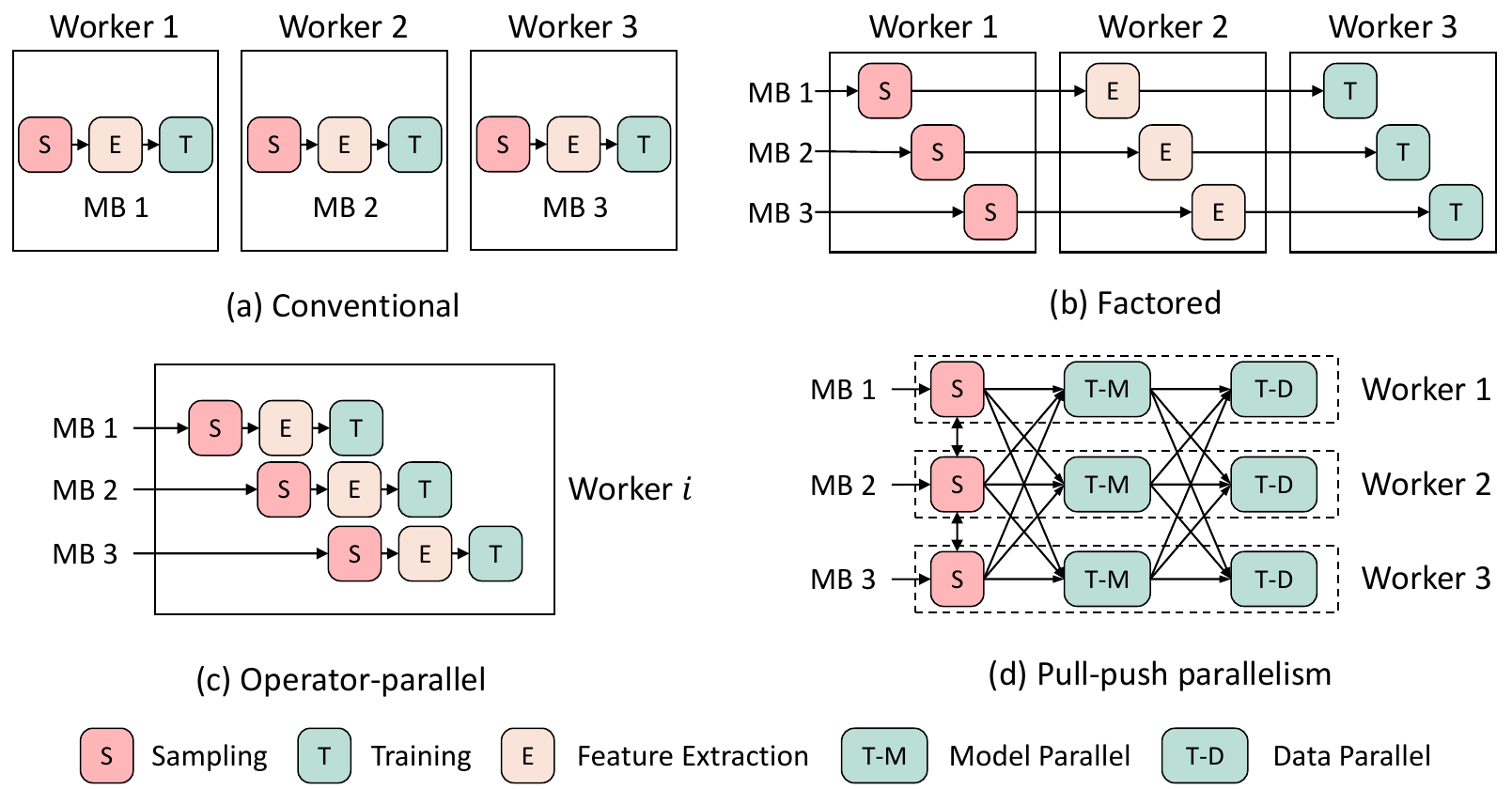}
    }
    \caption{\blue{Comparison of mini-batch execution models.}}
    \label{fig:mini-batch-exe-model}
\vspace{-20pt}
\end{figure}

\blue{\textbf{Conventional execution model} performs the sampling and feature extraction of each computation graph (i.e. batch) in sequence before batch training is performed, as shown in Figure~\ref{fig:mini-batch-exe-model}(a).}
When CPU-based graph sampling is applied, the sampling is executed on a CPU to construct the graph structure, feature extraction with respect to the graph is executed on a GPU with cache support, and training is executed on the GPU as well. When GPU-based sampling~\cite{nextdoor_eurosys_2021, csaw_sc_2020} is applied, all the operations in a pipeline are executed on a single device. Such conventional execution model is popular among existing GNN training systems~\cite{distdgl_ai3_2020, aligraph_vldb_2019, pagraph_socc_2020}. The optimizations in the context of the conventional execution model lie in the optimization of sampling operation and the cache strategy for feature extraction, as introduced in Section~\ref{sec:data-partition} and~\ref{sec:batch-generation}. However, the conventional execution model brings in resource contention. For example, processing graph sampling requests and subgraph construction compete for CPUs when CPU-based graph sampling is applied; graph sampling and feature extraction compete for GPUs' memory when GPU-based graph sampling is used.

\blue{\textbf{Factored execution model} uses a dedicated device or isolated resources to execute a single operation, as shown in Figure~\ref{fig:mini-batch-exe-model}(b). }
Empirical studies demonstrate that during the mini-batch GNN training, an operation shares a large amount of data in different epochs and preserves inter-epoch data locality. This observation motivates the factored execution model to avoid resource contention and improves the data locality.
Yang et al.~\cite{gnnlab} introduce a factored execution model that eliminates resource contention on GPU memory. The new model assigns a sampling operator and training operator to individual devices. More specifically, it uses some dedicated GPUs for graph sampling, and the remained GPUs are used for feature extraction and model training. With the help of the factored design, the new execution model is able to sample larger graphs while GPUs for feature extraction can cache more features. To improve the CPU utilization when CPU-based graph sampling is used. 
Liu et al.~\cite{bgl_nsdi23} assign isolated resources (i.e., CPUs) to different operations and balance the execution time of each operation by adjusting resource allocation. They introduced a profiling-based resource allocation method to improve CPU utilization and processing efficiency. The new resource allocation method first profiles the cost of each operator and solves an optimization problem minimizing the maximal completion time of all operations with resource constraints. 

\blue{\textbf{Operator-parallel execution model} enables the inter-batch parallelism and generates multi computation graphs in parallel by mixing the execution of all operators in the pipeline, as shown in Figure~\ref{fig:mini-batch-exe-model}(c). }
\blue{Zheng et al.~\cite{distdglv2_kdd_2022} decompose the batch generation stage into multiple stages (e.g., graph sampling, neighborhood subgraph construction, feature extraction) and each stage is treated as an operator.} Dependencies exist among different operators, while there are no dependencies among mini-batches. Therefore, the operators in different mini-batches can be scheduled into a pipeline, and operators without dependencies can be executed in parallel. 
\blue{In this way, the batch generation of a later batch can be performed immediately after or concurrently with that of the former batch}, instead of waiting for the former batch to finish the model training stage. 
\blue{
The idea introduced by Cai et al.~\cite{dsp-ppopp23} is very similar, as they set a sampler and a loader to conduct the sampling and feature extraction operation, respectively. A queue is used to manage the execution of different mini-batches. With a followed trainer to perform the execution of computation graph, the training process of mini-batches forms a training pipeline. Several mini-batches can be executed simultaneously, thus improving the GPU utilization and accelerating both the computation graph generation and execution stage. 
}
\blue{Zhang et al.~\cite{bytegnn_vldb_2022} also decouple the batch generation,} while further modeling the dependencies among different operators into a DAG. 
Treating each operator as a node in the DAG, a two-level scheduling strategy is proposed by considering both the scheduling among DAGs (coarse-grained) and among operators (fine-grained). 
With such two-level scheduling, the cost of CPU-based sampling and GPU-based feature extraction can be balanced, to improve the CPU utilization and reduce the end-to-end GNN training time.


\textbf{Execution model with pull-push parallelism}. All of the above execution models construct the subgraph with sampling first followed by extracting features according to the subgraph. When the input features are high-dimensional, the communication caused by the feature extraction becomes the bottleneck in the end-to-end GNN training pipeline. The expensive high-quality graph partition introduced in Section~\ref{sec:gp} is an option to reduce communication.  Gandhi et al.~\cite{p3_osdi_21} introduce a new execution model with pull-push parallelism. The new model partitions the graph using fast hash partition and the feature matrix using column-wise hash partition. 
It constructs the subgraph by pulling remote vertices and edges at local, and then pushes the constructed subgraph to all the workers. 
\blue{Then each worker extracts partial features and performs aggregation of these partial features for all the received subgraphs in parallel, forming the model parallel stage shown in Figure~\ref{fig:mini-batch-exe-model}(d). A data parallel stage which trains subgraphs (i.e. mini-batches) individually on each worker is then performed. Light communication overhead is incurred between the model parallel stage and data parallel stage to move the low-dimensional hidden embeddings.} 
With the help of pull-push parallelism, the execution model replaces the expensive input feature movement with the light graph structure movement, thus improving the efficiency of batch generation and model training.

\subsection{Full-graph execution model}
\blue{
For the distributed mini-batch GNN training, it is trivial to parallelize the execution of computation graphs, i.e., mini-batches, among the distributed workers, since each mini-batch is completely stored in a single worker. For the distributed full-graph or large-batch GNN training, the execution of a computation graph is partitioned among a set of workers. Due to the data dependency in the computation graph, it is non-trivial to achieve the high efficiency of the GNN training. So far, many computation graph execution models specific to GNN training have been proposed.} 

\blue{
In the following section, we first discuss the full-graph execution model from the perspective of  {vertex} 
computation. We delve into the execution model from both the view of graph operation (Section~\ref{sec:graphview}) and the matrix multiplication (Sectcion~\ref{sec:matrixview}). From the view of graph operation, the GNN execution can be modeled as SAGA-NN~\cite{neugraph_atc_2019}, which is inspired by the traditional GAS model in graph processing. From the view of matrix multiplication, the core of GNN execution can be modeled as SpMM (Sparse Matrix Multiplication). We also make a comparison in Section \ref{sec:graph-matrix-compare}. 
Second, we focus on the update mode, which determines whether the vertex embeddings and model parameters used for computation are updated in time or with delay. We categorize the full-graph execution model into synchronous execution models (Section~\ref{sec:sync-data-parallel}) and asynchronous execution models (Section~\ref{sec:async-data-parallel}). 
Note that the above two perspectives (i.e., vertex computation and update mode) are intersected, and a system simultaneously adopts one execution model from each of the two perspectives.
}

\subsubsection{Graph View.}
\label{sec:graphview}
\blue{
From the view of graph processing, we use the most well-known programming model \textit{SAGA-NN}~\cite{neugraph_atc_2019} for the following discussion. 
\textit{SAGA-NN} divides the forward computation of a single GNN layer into four operators \textit{Scatter} (SC), \textit{ApplyEdge} (AE), \textit{Gather} (GA) and \textit{ApplyVertex} (AV). SC and GA are two graph operations, in which vertex features are scattered along the edges and gathered to the target vertex, respectively. AE and AV may contain neural network (NN) operations, which process directly on the edge features or the aggregated features of the target vertices, respectively. 
}

According to different computation paradigms of graph operators (i.e., SC and GA), we divide computation graph execution models into \textit{one-shot execution} and \textit{chunk-based execution}.
\begin{figure}[!h]
    \centering
    \scalebox{0.33}{
    \includegraphics{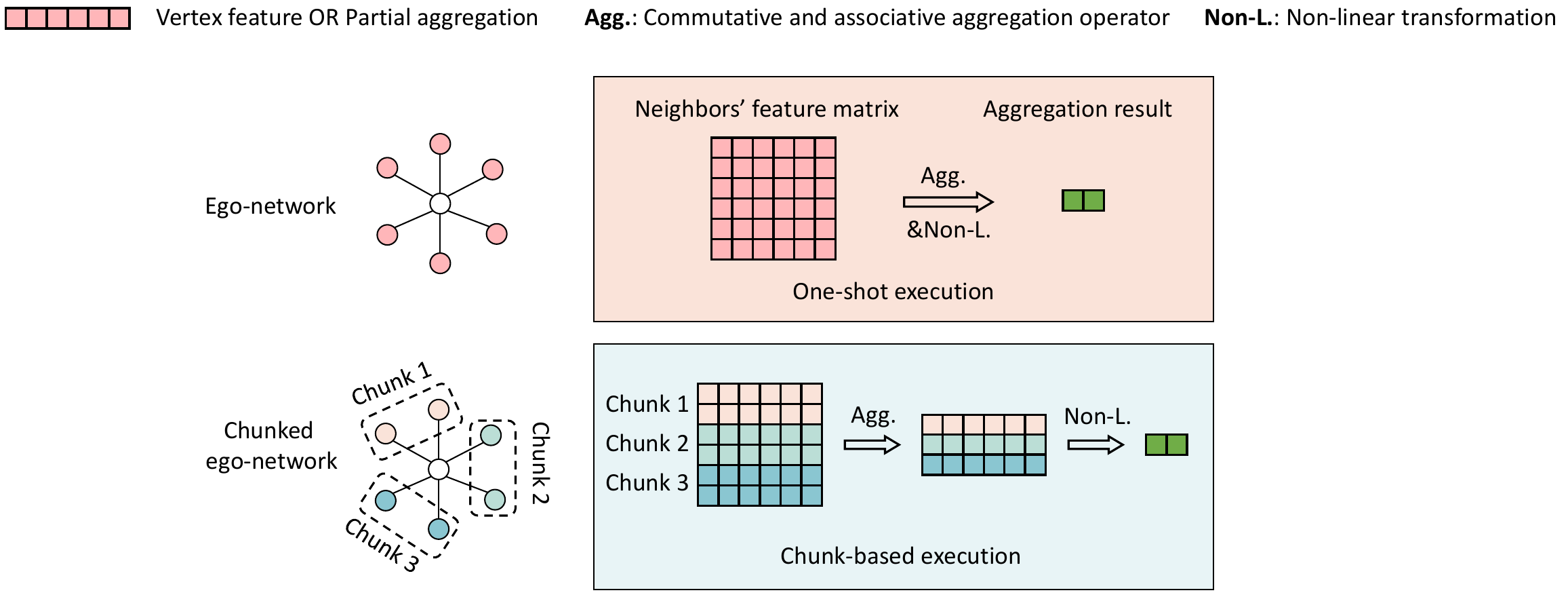}
    }
    \caption{One-shot vs. chunk-based execution model}
    \label{fig:one-shot-chunk}
    \vspace{-15pt}
\end{figure}

\textbf{One-shot Execution. } \label{sec:one-shot}
One-shot execution is the most common model adopted by GNN training. On each worker, all the neighbor features (or embeddings) needed for one graph aggregation of local vertices is collected in advance to the local storage, and the graph aggregation is processed in one shot, which is illustrated in the upper part of Figure~\ref{fig:one-shot-chunk}. The remote 
information may require communication with remote workers, and the communication protocol will be discussed in Section \ref{sec:communication-protocol}.
From the local view of graph aggregation, each target vertex is assigned to a specific worker. The worker gathers the neighbors' features of the target vertex and aggregates these features at a time.
As an example, DistDGL~\cite{distdgl_ai3_2020} adopts this model straightforwardly. BNS-GCN~\cite{BNSGCN} leverages a sampling-based method to aggregate neighbor vertices' features at random.
The one-shot execution can also be easily incorporated with the asynchronous execution model~\cite{wan2022pipegcn} (see Section~\ref{sec:async-data-parallel}), in which the stale embeddings are collected for the graph aggregation.   


\textbf{Chunk-based Execution.} \label{sec:chunk-based-execution}
The one-shot execution is straightforward, yet it causes expensive memory consumption and heavy network communication. For example, storing all the data required in the local storage of a worker may cause Out-Of-Memory (OOM) problem. On the other hand, transferring all the vertex features may lead to heavy network overhead. To address these problems, an alternative approach is to split the complete aggregation into several partial aggregations, and accumulate the partial aggregations to get the final graph aggregation. 
To achieve this, the neighborhoods of a vertex are grouped into several sub-neighborhoods, and we call each sub-neighborhood along with the vertex features or embeddings a \textit{chunk}. The chunks can be processed sequentially (\textit{sequential chunk-based execution}) or in parallel (\textit{parallel chunk-based execution}), which is illustrated in the lower part of Figure~\ref{fig:one-shot-chunk}.

Under the \textit{sequential chunk-based execution}, the partial aggregation is conducted sequentially and accumulated to the final aggregation result one after another. 
NeuGraph~\cite{neugraph_atc_2019} uses 2D partitioning method to generate several edge chunks, thus the neighborhood of a vertex is partitioned accordingly into several sub-neighborhoods. It
assigns each worker with the aggregation job of certain vertices, and feeds their edge chunks sequentially to compute the final result.
SAR~\cite{sar_mlsys22} uses edge-cut partitioning method to create chunks, retrieves the chunks of vertex from remote workers sequentially, and computes the partial aggregations at local. 
The sequential chunk-based execution effectively addresses the OOM problem, since each worker only needs to handle the storage and computation of one chunk at a time.

Under the \textit{parallel chunk-based execution}, the partial aggregations of different chunks are computed in parallel. After all chunks finish partial aggregation, communication is invoked to transfer the results, and the final aggregation result is computed at a time. 
Since the communication volume incurred by transferring the partial aggregation result is much less than transferring the complete chunk, the network communication overhead can be reduced. 
DeepGalois~\cite{deepgalois_mlsys21} straightforwardly adopts this execution model. DistGNN~\cite{distgnn_sc_2021} orchestrates the parallel chunk-based execution model with the asynchronous execution model, to transfer the partial aggregation with staleness. FlexGraph~\cite{flexgraph_eurosys_2021} further overlaps the communication of remote partial aggregations with the computation of the local partial aggregation to improve efficiency. 

\subsubsection{Matrix View.}
\label{sec:matrixview}

\newcommand{\StrategyA}{{partition strategy}\xspace}
\blue{
As described in Section \ref{sec:preliminary}, the matrix formulation of a GNN model is given by $\bm{H^l} = \sigma(\tilde{\bm{A}}\bm{H^{l-1}W^{l-1}})$, which involves SpMM since matrix $\bm{A}$ is sparse. In order to perform computations for the GNN model, three matrices (i.e., $\bm{A}$, $\bm{H}$, and $\bm{W}$) need to be stored either locally or in a distributed manner. For GNN on large graphs, it is impractical for a single worker (e.g., a GPU processor) to store all three matrices simultaneously. At least one matrix must be partitioned and distributed across different workers. 
The execution of distributed SpMM in GNN can be divided into three stages: communication, computation, and reduction. 
The computation stage is the core of matrix multiplication, which is performed locally in each worker.
Prior to the computation stage, workers may require specific blocks of a matrix from other workers, necessitating a communication stage. Typically, the communication stage is accomplished through a broadcasting mechanism.
After the computation stage, a worker may only possess partial results of the final matrix block. In such cases, a reduction stage is necessary to gather the remaining partial results from other workers. Workers retrieve and sum up these partial results to obtain the final matrix.}

\blue{According to the existence of the above three stages, we categorize the distributed SpMM into three execution models: \textit{computation-only}, \textit{communication-computation}, and \textit{communication-computation-reduction}. 
The existence of the communication and reduction stage is determined by two factors: \ding{192}\textbf{\StrategyA}: how the matrix is partitioned and stored; \ding{193}\textbf{stationary strategy}: which matrix is kept as the \blue{stationary matrix (i.e., no communication is incurred to move data in this matrix)}. The \StrategyA determines whether a matrix is replicated stored on multiple workers, or partitioned into blocks and distributed to different workers. The \StrategyA also dictates the matrix partitioning mechanisms, such as 1D, 2D, etc. The stationary strategy determines the choice of the stationary matrix. During the execution of distributed SpMM, each partitioned block of the stationary matrix is pinned on the corresponding worker, and no communication is required for the stationary matrix. We summarize how these factors impact the execution model in Table~\ref{tab:spmm-exe-model}, which will be introduced in detail in the following. 
}

\blue{As discussed in Section \ref{sec:preliminary}, the weight parameter matrix $\bm{W}$ is relatively small in GNN models. As a result, $\bm{W}$ is fully replicated across all workers, adhering to the basic data parallelism principle. 
The subsequent discussion primarily focuses on \blue{matrices $\bm{A}$, $\bm{H}$, and their product $\bm{P} = \bm{A}\bm{H}$}.
}


\blue{
\textbf{Computation-only Execution Model.} In this model, both the communication and reduction stages are eliminated by adopting specific \StrategyA. One of $\bm{A}$ or $\bm{H}$ is fully replicated across each worker, while the other matrix is partitioned properly among different workers to achieve a communication-free paradigm~\cite{Redistribution-RDM}. For instance, matrix $\bm{A}$ is replicated across all workers, and matrix $\bm{H}$ is partitioned into column blocks. Each worker holds a column block, which contains multiple columns of $\bm{H}$. In such case, no communication is required prior to local computation. After local computation, each worker holds a corresponding column block of the final matrix $\bm{P}=\bm{A}\bm{H}$, thus no reduction is required either. In this case, only the computation stage is performed in the SpMM operation.
However, this execution model lacks strong scalability if both matrices $\bm{A}$ and $\bm{H}$ exceed the memory capacity of an individual worker, since at least one of the matrices needs to be fully replicated across all workers.
}


\newcommand{\aStat}{{$\bm{A}$-Stationary}\xspace}
\newcommand{\hStat}{{$\bm{H}$-Stationary}\xspace}
\newcommand{\pStat}{{$\bm{P}$-Stationary}\xspace}
\blue{
\textbf{Communication-computation Execution Model.} 
For the circumstances that both matrix $\bm{A}$ and $\bm{H}$ have to be partitioned and stored on the workers in a distributed manner, the communication-computation execution model is introduced. In this execution model, the workers need to share the matrix partitions they hold with each other, necessitating the communication stage prior to local computation. This communication can be performed either in a broadcast fashion or a point-to-point (P2P) fashion, as detailed in Section~\ref{sec:communication-protocol}. 
Furthermore, since the reduction stage is not executed, $\bm{P}$-Stationary strategy is adopted in the communication-computation execution model and no communication is required to obtain the result matrix $\bm{P}$.}

\blue{The third column in Table~\ref{tab:spmm-exe-model} shows that when \pStat strategy is adopted, many partition strategies (1D, 1.5D and 2D) can be applied in the communication-computation execution model.
For the \pStat 1D partitioning, each worker stores a row block of matrices $\bm{A}$, $\bm{H}$, and $\bm{P}$. During the communication stage, each worker broadcasts its row block of $\bm{H}$ to all other workers. Subsequently, local computation is performed to compute the block rows of $\bm{P}$. 
In this paradigm, matrix $\bm{A}$ is also stationary, making this 1D \pStat SpMM also \aStat. 
It is worth noting that the 1D partitioning can also be performed in a column-wise manner~\cite{p3_osdi_21}. In such cases, the 1D \pStat is also \hStat. 
In short, under 1D partitioning, both \aStat and \hStat are equivalent to \pStat. Therefore, the 1D partitioning follows a communication-computation execution model.}
\blue{However, 1D \pStat faces scalability challenges as a {worker} needs to broadcast the partition to all remote workers, resulting in linear communication costs proportional to the number of workers~\cite{partition_sc_2020}. To address this, optimizations such as P2P communication and non-blocking techniques~\cite{spmm-vldb23, sancus_vldb_2022} can be employed to accelerate the communication stage, {which will be discussed in Section~\ref{sec:communication-protocol}.}}

\blue{For \pStat 1.5D partitioning, either matrix $\bm{A}$ or matrix $\bm{H}$ is partitioned in a 2D manner, while the other matrix is partitioned in a 1D manner. To partition a matrix in a 2D manner, each processor holds a row-column block of the complete matrix, comprising elements in the matrix that satisfy both the assigned column IDs and row IDs for the processor. Under 1.5D partitioning, although matrix $\bm{P}$ is set to be stationary, either matrix $\bm{A}$ or matrix $\bm{H}$ needs to be broadcast to all processors, leading to scalability challenges similar to 1D partitioning. }

\blue{Another approach is to leverage \pStat 2D partitioning, where both matrix $\bm{A}$ and matrix $\bm{H}$ are partitioned in a 2D manner. 
For a row-column block in matrix $\bm{P}$, the processor holding this block also holds the corresponding row-column blocks for matrix $\bm{A}$ and matrix $\bm{H}$. It only needs to receive the blocks with the same row ID of matrix $\bm{A}$ and the blocks with the same column ID of matrix $\bm{H}$. In this way, the total communication overhead is reduced.
}

\blue{
\textbf{Communication-computation-reduction Execution Model.}
As described above, adopting \pStat strategy is the key to eliminating the reduction stage. Therefore, in this model, \pStat strategy is not adopted. In other words, different from the communication-computation execution model where \pStat strategy is adopted, other stationary strategies including \aStat, \hStat, and Non-Stationary strategies are considered. 
As discussed previously, under both replicated and 1D partitioning, all stationary strategies are equivalent to \pStat, and no reduction stage is required. 
For \aStat and \hStat, under 1.5D and 2D partitioning, the results obtained after {the local computation stage are still partial}. Each worker performs a reduction operation to sum up these remote partial results with its local partial result.
The 3D partitioning can be viewed as the aggregation of multiple 2D partitioning SpMM operations. 
As a result, a final reduction operation is required to aggregate the results of multiple 2D partitioning SpMM operations. 
Under 3D partitioning, non of the three matrices $\bm{A}, \bm{H}, \bm{P}$ can be stationary. Therefore, 3D partitioning only corresponds to Non-Stationary strategy.
}

\begin{table}[]
\resizebox{0.6\textwidth}{!}{
    \begin{tabular}{ccccc}
    \hline
                        & \textbf{\aStat} & \textbf{\hStat} & \textbf{\pStat} & \textbf{Non-Stationary} \\ \hline
    \textbf{Replicated} & C               & C               & C               & -                       \\ \hline
    \textbf{1D}         & CC              & CC              & CC              & -                       \\ \hline
    \textbf{1.5D}       & CCR             & CCR             & CC              & -                       \\ \hline
    \textbf{2D}         & CCR             & CCR             & CC              & -                       \\ \hline
    \textbf{3D}         & -               & -               & -               & CCR                     \\ \hline
    \end{tabular}
}
\caption{\blue{Two factors in distributed SpMM execution model. The rows indicates the \StrategyA, and columns indicates the stationary strategy. \textbf{C}: Computation-only. \textbf{CC}: Communication-computation. \textbf{CCR}: Communication-computation-reduction. The Non-Stationary corresponds to 3D partition, where none of the three matrix is kept stationary during the execution.}}
\label{tab:spmm-exe-model}
\vspace{-35pt}
\end{table}

\blue{
\textbf{Discussion.}
Both \StrategyA and stationary strategy play crucial roles in distributed SpMM computation. We summarize the impact of these two factors on the distributed SpMM execution model as shown in Table~\ref{tab:spmm-exe-model}. 
For the circumstances that one of $\bm{A}$ and $\bm{H}$ is replicated stored, a communication-free SpMM can be performed, leading to a computation-only execution model. However, since the GNN model consists of both SpMM and GeMM (General dense Matrix Multiplication) operations, a redistribution step is required between them to achieve both communication-free SpMM and GeMM~\cite{Redistribution-RDM}. Additionally, the order in which SpMM and GeMM are executed (i.e., SpMM first or GeMM first) can also impact the communication and computation overhead~\cite{Redistribution-RDM, GNNAdvisor_OSDI_2021}.
When both matrices $\bm{A}$ and $\bm{H}$ are distributed across different processors, and the 1D partitioning method or the \pStat is adopted, SpMM follows the \textit{communication-computation} execution model. However, this execution model faces scalability and communication overhead.
Finally, for \aStat or \hStat with 1.5D or 2D partitioning, and for the Non-Stationary strategy, they follow the \textit{communication-computation-reduction} execution model. The communication cost in the overall SpMM operation varies depending on the selected stationary mechanism, and different matrix partitioning methods contribute differently to the communication cost, as discussed in more detail in previous studies~\cite{partition_sc_2020, tall-skinny-ics21}.
}

\subsubsection{Comparison of graph view and matrix view. } \label{sec:graph-matrix-compare}
\blue{
We provide a comparison between the graph view and matrix view of GNN models and highlight the connections between these two views. The graph view and matrix view are two different programming interfaces that handle GNN models at different levels. The matrix view is a lower-level representation of GNN models, as it directly manages the 
data presented in matrices. On the other hand, the graph view provides a higher-level understanding of GNN models. In most GNN frameworks such as DGL and PyG, the programming interfaces are organized based on the graph view for ease of use and convenience.
However, during model training, graph operations such as feature aggregation are ultimately transformed into matrix multiplications. Both the one-shot execution and chunk-based execution in the graph view can be mapped to the matrix view. In a simple one-shot execution, each vertex is assigned to a specific worker along with its input features. The worker performs an aggregation when it receives the features of all its neighbors. This corresponds to the 1D partitioning $\bm{P}$-Stationary in the matrix view, which follows a \textit{communication-computation} paradigm. In other stationary mechanisms that follow a \textit{communication-computation-reduction} paradigm, the partial results obtained from the reduction operation correspond to the partial aggregation results in the chunk-based execution from the graph view.
}

\subsubsection{Synchronous execution model} \label{sec:sync-data-parallel}
In synchronous execution model (Figure~\ref{fig:sync}), the computation of each operator begins after the communication is synchronized. 
\blue{
Almost all distributed SpMM executions follow a synchronous model. Therefore, to achieve a better understanding of the difference between synchronous and asynchronous execution models while also adhering to the page limit, we introduce the following from the graph view only with the SAGA-NN model as an example. 
}
In SAGA-NN model, four operators and their backward counterparts form an execution pipeline for the graph data, following the order of SC-AE-GA-AV-$\triangledown$AV-$\triangledown$GA-$\triangledown$AE-$\triangledown$SC. 
{Among these eight stages, two of them involve the communication of states of boundary vertices, that is, GA and $\triangledown$GA. In GA, neighbor vertices features should be aggregated to the target vertices, thus the features of boundary vertices should be transferred. In $\triangledown$GA, the gradients of the boundary vertices should be sent back to their belonging workers. Therefore, GA and $\triangledown$GA are two synchronization points in synchronous execution model.
On these points, the execution flow has to be blocked until all communication finishes. 
}
Systems like {NeuGraph~\cite{neugraph_atc_2019}}, CAGNET~\cite{partition_sc_2020}, FlexGraph~\cite{flexgraph_eurosys_2021}, DistGNN~\cite{distgnn_sc_2021} apply this execution model. 
To reduce the communication cost and improve the training efficiency, several communication protocols are proposed and we review them in Section~\ref{sec:sync_comm_protocol}. 
\blue{We point out that these two synchronization points exist irrespective of the view of execution model. From a matrix view, GA corresponds to the $AH$ SpMM operation, and $\triangledown$GA corresponds to the $AG$ SpMM operation, where $G$ represents the gradient matrix in backward propagation.}


\subsubsection{Asynchronous execution model} \label{sec:async-data-parallel}
The asynchronous execution model allows the computation starts with historical states and avoids the expensive synchronization cost. According to the different types of states, we classify the asynchronous execution model into \textit{type I asynchronization} and \textit{type II asynchronization}.
\begin{figure}[t]
\subfigure[Synchronous execution model]{
\scalebox{0.25}{
   \includegraphics{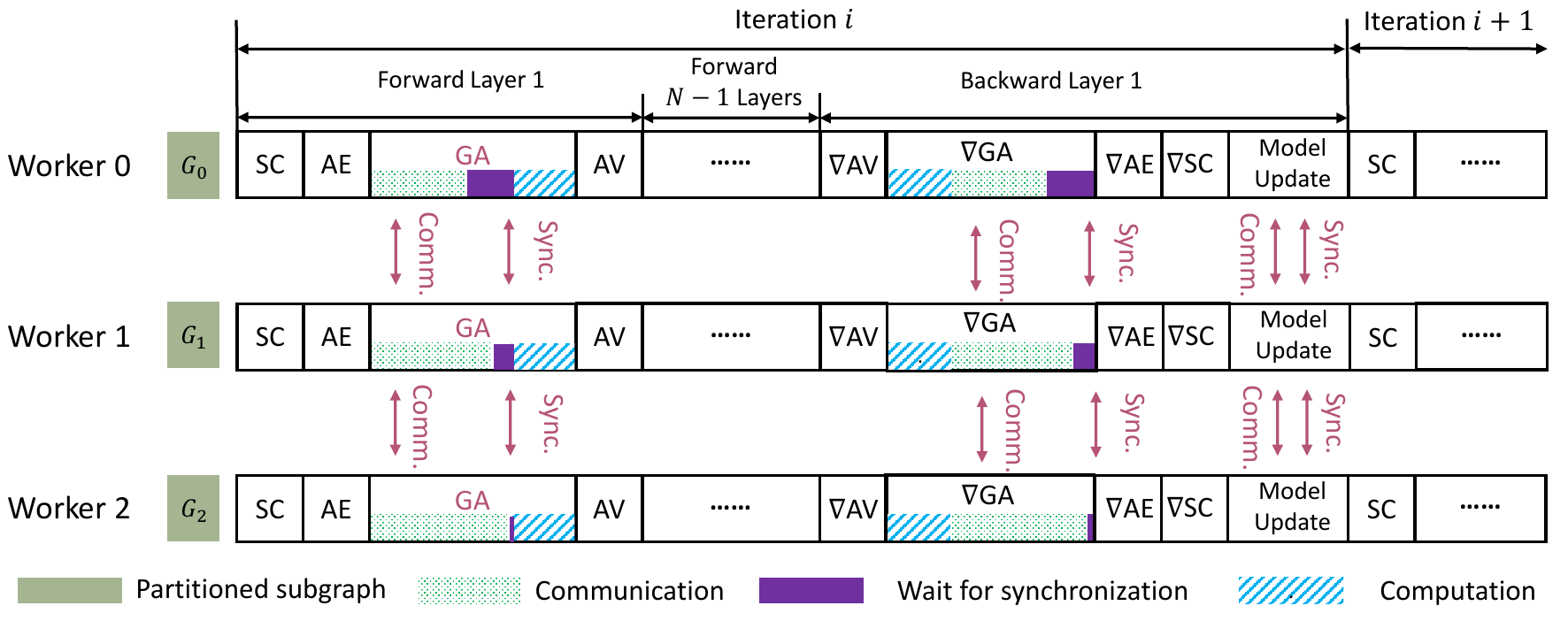}
}
   \label{fig:sync}
}
\subfigure[Asynchronous execution model (Type I)]{
\scalebox{0.23}{
   \includegraphics{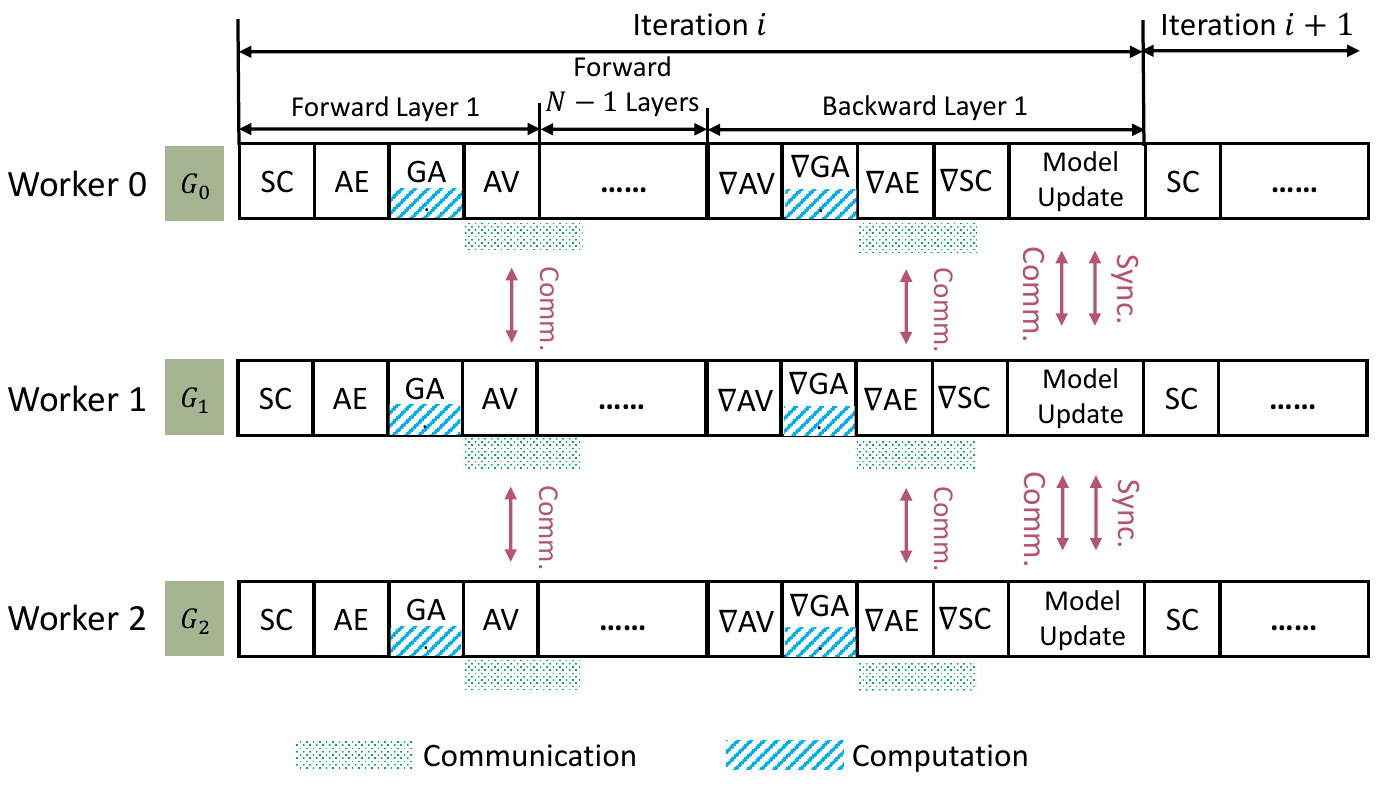}}
   \label{fig:async1}
}
\vspace{-10pt}
\caption{\blue{Synchronous and Asynchronous execution model.}}
\label{fig:execution_models}
\vspace{-10pt}
\end{figure}

\textbf{\textit{Type I asynchronization}}. {As mentioned above, two synchronization points exist in the execution pipeline of the SAGA-NN model, and similar synchronization points exist in other programming models as well. 
Removing such synchronization points of computation graph computation in the execution pipeline introduces \textit{type I asynchronization}~\cite{sancus_vldb_2022, distgnn_sc_2021, wan2022pipegcn}. In type I asynchronization (Figure~\ref{fig:async1}), workers do not wait for the hidden embeddings (hidden features after the first GNN layer) of boundary vertices to arrive. Instead, it uses the historical hidden embeddings of the boundary vertices from previous epochs which are cached or received earlier to perform the aggregation of target vertices. Similar historical gradients are used in the $\triangledown$GA stage. 
}

\textbf{\textit{Type II asynchronization}}. {
\blue{During GNN training, another synchronization point occurs when the weight parameters need to be updated. This is identical to traditional deep learning models, where many efforts have been made to remove such synchronization point~\cite{hogwild_arxiv_2011,pipedream_SOSP_2019}.
Gandhi et al.~\cite{p3_osdi_21} further adopt this idea into GNN models, which forms \textit{type II asynchronization}.
Under such protocol, mini-batches are executed in parallel, and form a mini-batch pipeline.
} 
}

\blue{
Note that type I and type II asynchronization can be adopted individually or jointly. 
Detailed communication protocol for GNN-specific Type I asynchronous execution model is described in Section \ref{sec:async-aggregation}.
}

\section{GNN Communication Protocols}\label{sec:communication-protocol}

In this section, we review GNN communication protocols that support different computation graph execution models, and they are responsible to transfer hidden embeddings and gradients. 
According to the types of computation graph execution models, we classify GNN communication protocols into synchronous GNN communication protocols and asynchronous GNN communication protocols.  

\subsection{Synchronous GNN Communication Protocols}
\label{sec:sync_comm_protocol}
\blue{Synchronous GNN communication protocols handle the communication in a synchronous execution model, where a synchronous barrier is set in GA and $\triangledown$GA of each layer, to ensure the hidden embeddings used for aggregation are computed from exact last layer in the current iteration.
There are several approaches to transferring the hidden embeddings, and we introduce them one by one as follows.
}


\subsubsection{Broadcast-based protocol}
Broadcast is a general technique to share information among a set of workers. 
In the GNN context, each worker broadcasts the latest hidden embeddings or graph structure it holds according to the underlying data partitioning methods, and other workers start the corresponding graph convolutional computation once they receive the required data from that worker. When all workers finish broadcasting, the hidden embeddings of the full graph are shared among all workers, and all workers receive the latest hidden embeddings they need in the current iteration. 
The authors of CAGNET~\cite{partition_sc_2020} model the GNN computation as matrix computation. They partition the matrices (i.e., graph adjacency matrix and feature matrix) into sub-matrices with 1-D, 1.5-D, 2-D, and 3-D graph partitioning strategies. During the GNN training, each worker directly broadcasts the partial matrix of graph structure or vertex features to other workers for synchronization, guaranteeing that the vertices on a worker have a complete neighborhood.
The efficiency of the broadcast can be destroyed by the workload imbalance problem, which causes a stall in computation.

\subsubsection{Point-to-point-based protocol}
\label{sec:ptop-protocol}
The aforementioned broadcast sends data to all the workers in the distributed environment and might cause redundant communication. In contrast, Point-to-Point (P2P) transmission is a fine-grained communication method to share information among workers, and it only transfers data alongside cross edges in an edge-cut graph partition or replicated vertices in a vertex-cut graph partition. 
\blue{For example, ParallelGCN~\cite{spmm-vldb23} leverages the edge-cut graph partitioning, and maintains a special data structure to manage the different vertex sets for different worker pairs. Each vertex set indicates the vertices that need to be sent between the two workers. The communication of these vertices is then performed in a P2P fashion. The communication volume is thus reduced since only necessary vertices are sent and received.}
DistGNN~\cite{distgnn_sc_2021} employs the vertex-cut partitioning strategy. Each edge is assigned to one specific worker, while each vertex along with its features may be stored repetitively. Among the replications, one vertex is called the master vertex. When aggregating the neighbors of a vertex, P2P transmission is applied. Specifically, partial aggregations of the replications of the master vertex are sent to the worker where the master vertex is located, instead of broadcasting to all workers. As a result, the master vertex can aggregate the embeddings from all its neighbors. 
However, according to the studies of CAGNET~\cite{partition_sc_2020}, the naive P2P transmission brings in the overhead of \textit{request and send} operations and increases the latency. 
To mitigate this extra overhead, pipeline-based protocol is introduced, which further overlaps the communication with local computation, as described in Section~\ref{sec:pipeline-based-protocol}.

\subsubsection{Pipeline-based protocol}\label{sec:pipeline-based-protocol}
\blue{Since the naive P2P-based protocol leads to an extra overhead of \textit{request and send} operations introduced in Section~\ref{sec:ptop-protocol}, pipeline-based protocol is proposed as an optimization technique. It divides the complete task of vertex aggregation on workers to several sub-tasks, and overlap the communication and computation of these sub-tasks. The execution of these sub-tasks thus form a pipeline.}

\blue{G3~\cite{g3-sigmod23} introduces a bin packing mechanism to pack local vertices into different bins (i.e., sub-tasks). During the aggregation stage, some vertices may be ready to execute the aggregation of the next layer prior to other vertices once all of their remote neighbors' embeddings are received. These vertices are prioritized to be packed into bins and compute the aggregation results. Once the computation of a bin finishes, the aggregation result of the vertices in that bin is sent to remote workers. In this way, a fine-grained overlap of communication and computation can be achieved, and different bins form a pipeline.  
}

\blue{
In systems that adopt the chunk-based execution model introduced in Section~\ref{sec:chunk-based-execution}, such as SAR~\cite{sar_mlsys22}, DistGNN~\cite{distgnn_sc_2021} and FlexGraph~\cite{flexgraph_eurosys_2021}, the aggregation of each neighborhood chunk can be treated as a sub-task, providing the opportunity to overlap the communication and computation of different chunks. 
}
Here we call the aggregation of a vertex's partial neighborhood the \textit{partial aggregation}. 
SAR~\cite{sar_mlsys22} aggregates the chunks in a predefined order. Each worker in SAR first computes the partial aggregation of the local neighborhood of the target vertex, then fetches the remote neighborhood in the predefined order, computes the partial aggregation of the remote neighborhood, and accumulates the results at local one-by-one.
Consequently, the aggregation of the complete neighborhood in SAR is divided into different stages, and in each stage, a partial aggregation is performed. The final aggregation is computed when all partial aggregation completes. SAR applies the pipeline-based communication protocol to reduce memory consumption since it does not need to store all the features of the neighborhood and compute the aggregation at once. 
\blue{ParallelGCN adopts a similar idea without pre-defining the aggregation order. The remote chunks of neighborhood are received in a random order, and it starts aggregating the received vertices immediately.} 
Furthermore, to reduce the overhead of network communication caused by transferring of the partial neighborhood, DeepGalois~\cite{deepgalois_mlsys21} performs a remote partial aggregation before communication. Therefore, only the partial aggregation results need to be transferred. The cd-0 communication strategy in DistGNN is similar, which simply performs remote partial aggregation, fetches it to the local worker, and computes the final aggregation result.
To further improve the training efficiency, FlexGraph~\cite{flexgraph_eurosys_2021} overlaps the computation and communication. Each worker first issues a request to partially aggregate the neighborhood at the remote worker and then computes the local partial aggregation while waiting for the remote partial aggregations to complete the transfer. After receiving the remote partial aggregations, the worker directly aggregates them with the local partial aggregation. Therefore, the local partial aggregation is overlapped with the communication to get remote partial aggregation.

\subsubsection{Communication via Shared Memory}
To train large GNN at scale, it is also feasible to leverage the CPU memory to retrieve the required information instead of GPU-GPU communication. The complete graph and feature embeddings are stored in shared memory (i.e., CPU memory) and the device memory of each GPU is treated as a cache. In ROC~\cite{roc_mlsys_2020}, the authors assume that all the GNN data is fit in the CPU memory and they repetitively store the whole graph structure and features in the CPU DRAM on each worker. GPU-based workers retrieve the vertex features and hidden embeddings from the local CPU-shared memory. For larger graphs or scenarios that all the GNN data cannot fit into the CPU DRAM of a single worker, distributed shared memory is preferred. DistDGL~\cite{distdgl_ai3_2020} partitions the graph and stores the partitions with a specially designed KVStore. During the training, if the data is co-located with the worker, DistDGL accesses them via local CPU memory, otherwise, it incurs RPC requests to retrieve information from other remote CPU memory.
{
In NeuGraph~\cite{neugraph_atc_2019}, the graph structure data and hidden embeddings are also stored in shared memory. To support large-scale graphs, the input graph is partitioned into $P\times P$ edge chunks by the 2D partitioning method, and the feature matrix (as well as hidden embedding matrix) is partitioned into $\bm{P}$ vertex chunks by the row-wise partition method. As described in Section \ref{sec:chunk-based-execution}, it follows a chunk-based execution model. To compute the partial aggregation of a chunk, each GPU  fetches the corresponding edge chunk as well as the vertex chunk from the CPU memory. To speed up the vertex chunk transfer among GPUs, a chain-based streaming scheduling method is applied considering the communication topology to avoid bandwidth contention.
}

\subsection{Asynchronous GNN communication protocol} \label{sec:async-aggregation}
Asynchronous GNN communication protocols are used to support asynchronous execution models.
As described in Section \ref{sec:async-data-parallel}, there are two asynchronization in GNN training, type I asynchronization which removes the synchronization in aggregation stages and type II asynchronization which removes synchronization in model weight updating stages. Type I asynchronization is specific to GNN model computation, while type II asynchronization has been widely used in traditional deep learning and the corresponding techniques can be directly {introduced into} GNN training. In the following, we focus on reviewing the asynchronous GNN communication protocols that are specific to distributed GNN training.
Table~\ref{tab:async-parallel} summarizes the popular asynchronous GNN communication protocols.~

\begin{table*}[t]
\centering
\caption{{\blue{Summary of asynchronous communication protocols. The staleness bound $s$ is set by users.}}}
\label{tab:async-parallel}
\resizebox{\textwidth}{!}{%
\begin{tabular}{ccccccccc}
\hline
                  & \textbf{Type I} & \textbf{Type II} & \tabincell{c}{\textbf{Use Stale}\\\textbf{Embedding}} & \tabincell{c}{\textbf{Use Stale}\\\textbf{Gradient}} & \textbf{Staleness Model} & \textbf{Staleness bound} & \textbf{Parallelism} & \textbf{Mechanism}                \\ \hline
\textbf{SANCUS}   & \cmark          & \xmark           & \cmark                       & \xmark                      & Variation-based          & Algorithm Depended       & Data                 & Skip-broadcast                    \\ \hline
\textbf{DIGEST}   & \cmark          & \xmark           & \cmark                       & \xmark                      & Epoch-adaptive           & s                        & Data                 & Embedding server                  \\ \hline
\textbf{DIGEST-A} & \cmark          & \cmark           & \cmark                       & \xmark                      & Epoch-adaptive           & s                        & Data                 & Embedding server                  \\ \hline
\textbf{DistGNN}  & \cmark          & -                & \cmark                       & \xmark                      & Epoch-fixed              & s                        & Data                 & Overlap                           \\ \hline
\textbf{PipeGCN}  & \cmark          & \xmark           & \cmark                       & \cmark                      & Epoch-fixed              & 1                        & Data                 & Pipeline                          \\ \hline
\textbf{Dorylus}  & \cmark          & \cmark           & \cmark                       & \cmark                      & Epoch-adaptive           & s                        & Data                 & Pipeline                          \\ \hline
\textbf{P3}       & \xmark          & \cmark           & \xmark                       & \xmark                      & Epoch-fixed              & 3                        & Model                & Mini-batch pipeline            \\ \hline \hline
\textbf{CM-GCN}   & -               &-                  & -                       & -                      & -                        & 0                        & Data                 & Counter-based vertex-level async. \\ \hline
\end{tabular}%
}
\end{table*}


\subsubsection{Staleness Models for Asynchronous GNN communication protocol}
{ 
Generally, bringing in asynchronization means information with staleness should be used in training. Specifically, in type I asynchronization, the GA or $\triangledown$GA stage is performed without completely gathering the latest states of the neighborhood, and the historical vertex embeddings or gradients are used in aggregation.
Different staleness models are introduced to maintain the staleness of historical information, and each of them should ensure that the staleness of aggregated information is bounded so that the GNN training is converged.} In the following, we review three popular staleness models.

{\textbf{Epoch-Fixed Staleness.} One straightforward method is to aggregate historical information with fixed staleness~\cite{wan2022pipegcn,distgnn_sc_2021,p3_osdi_21}. 
Let $e$ be the current epoch in training, and $\tilde{e}$ be the epoch of the historical information used in aggregation. In epoch-fixed staleness model, $|\tilde{e} - e| = \epsilon_E$, where $\epsilon_E$ is a hyper-parameter set by users. In this way, the staleness is bounded explicitly by $\epsilon_E$.}

{\textbf{Epoch-Adaptive Staleness.}
Another variation of the above basic model is to use an epoch-adaptive staleness~\cite{dorylus_osdi_2021,digest_arxiv_2022}. During the aggregation of a vertex $i$, let $\tilde{e}_{N_j}$ be the epoch of historical information of any vertex $j \in N(i)$ (i.e., any neighbor vertex of $i$) used in aggregation. In epoch-adaptive staleness model, for all $j \in N(i)$, $|\tilde{e}_{N_j} - e| \leqslant \epsilon_A$ holds. This means in different epochs, the staleness of historical information for aggregation may be different, and in one epoch, the staleness of aggregated neighbors of a vertex may also be different. Generally, once $\tilde{e} - e$ reaches $\epsilon_A$, then the latest embeddings or gradients should be broadcast under decentralized training or pushed to the historical embedding server under centralized training. If the above condition does not hold when aggregation is performed, then the aggregation is blocked until the historical information within the staleness bound (i.e., $\epsilon_A$) can be retrieved.
}

{\textbf{Variation-Based Staleness.}
Third, the staleness can also be measured by the variation of embeddings or gradients. In other words, we only aggregate embeddings or gradients when they are significantly changed. Specifically, 
let $H_w^{(l)}$ be the embeddings of layer $l$ held by worker $\bm{W}$, and $\tilde{H}_w^{(l)}$ be the historical embedding last shared to other workers. 
In the variation-based staleness model, $H_w^{(l)} - \tilde{H}_w^{(l)} \leqslant \epsilon_V$, where $\epsilon_V$ is the maximum difference bound set by users so that the latest embeddings or gradients are broadcast if the historical version available for other workers are too stale. In this way, the staleness is bounded with the help of $\epsilon_V$.
}

\subsubsection{Realization of Asynchronous GNN communication protocols}\label{sec:asyn-protocol-realization}

As described above, historical information can be used in aggregation during GA or $\triangledown$GA stages. Some protocols only take the GA stage into consideration, supporting asynchronous embedding aggregation. Some protocols are designed to use historical information during both  GA and $\triangledown$GA stages, supporting both asynchronous embedding and gradient aggregation. 

\textbf{Asynchronous Embedding Aggregation.}
Peng et al.~\cite{sancus_vldb_2022} take the basic training paradigm of broadcast in CAGNET into consideration, and design a skip-broadcast mechanism under a 1D partition. This mechanism auto-checks the staleness of the partitioned vertex features on each compute vertex and skips the broadcast if it is not too stale. If the broadcast of a partition is skipped, then other workers should use the historical vertex embeddings cached previously to process the forward computation. Three staleness check algorithms are further proposed, and each one of them ensures that the staleness of cached vertex embeddings is bounded. 

{Chai et al.~\cite{digest_arxiv_2022} adopt a similar idea while using parameter servers to maintain the historical embeddings of all workers. The vertex hidden embeddings are pushed to these servers every several epochs, and the workers pull the historical embeddings to their local cache in the next epoch after the push. The staleness of historical vertex embeddings is thus bounded by the push-pull period.}

The cd-r algorithms introduced by Md et al.~\cite{distgnn_sc_2021} overlap the communication of partial aggregation results from each worker with the forward computation in GNN. Specifically, the partial aggregation in iteration $i$ is transmitted asynchronously to the target vertex, and the final aggregation is performed in iteration $i + r$. Under this algorithm, the staleness is bounded to $r$.

\textbf{Asynchronous Embedding and Gradients Aggregation.}
As we introduced in Section \ref{sec:sync-data-parallel}, GA and $\triangledown$GA are two synchronization points in the GNN execution model. While the above protocols only use historical vertex embeddings during stage GA, Wan et al.~\cite{wan2022pipegcn} also take the synchronization point $\triangledown$GA into consideration, and use stale vertex gradients in the back propagation. {During the training, both embeddings and gradients from the neighborhood are asynchronously sent in a point-to-point fashion in the last epoch and received by the target vertex in the current epoch. Therefore, the communication is overlapped in both forward and backward computation and the training pipeline with a fixed-epoch gap is thus constructed}, in which workers are only allowed to use features or gradients exactly one epoch ago, and the staleness is bounded.
Thorpe et al.~\cite{dorylus_osdi_2021} design a finer and more flexible pipeline. 
Similarly to the method proposed by Wan et al.~\cite{wan2022pipegcn}, both stale vertex embeddings and vertex gradients are used in the pipeline.
Moreover, it also removes the type II synchronization, and thus a trainer may use stale weight parameters and starts the next epoch immediately. 
The staleness in the pipeline is explicitly bounded with $S$ set by users. The bounded staleness $S$ ensures that the trainer which moves the fastest in the pipeline is allowed to use historical embeddings or gradients at most $S$ epochs earlier. With this staleness bound, the staleness of weight parameters is also bounded accordingly.



\textbf{Vertex-level Asynchronization.}
{
Without using stale embeddings or gradients, asynchronization can also be designed in vertex-level~\cite{distgnn_CMGCN}, where each vertex starts the computation of the next layer as soon as all its neighbors' embeddings are received. Different from the traditional synchronous method in which all vertices on a worker start the computation of a layer together, during vertex-level asynchronous processing, different vertices in one worker may be computing different layers at the same time. Note that this asynchronization does not make any difference in the aggregation result, since all the information required should be the latest, and no embeddings or gradients from previous epochs are used. 
}

\section{Distributed GNN Training Systems}
\label{sec:system}

\begin{table*}[!h]
\caption{{Summary of \numsys distributed GNN training systems and frameworks.}}
\vspace{-10pt}
\label{tab:systems}
\centering
\resizebox{\textwidth}{!}{%
\begin{tabular}{c|c|c|c|c|c|c|c|c|c}
\hline
\textbf{Systems}                          & \textbf{\textbf{Hardware}}    & \textbf{\textbf{Cost model}} & \textbf{\textbf{Graph partition}}                                   & \textbf{\textbf{Feature Partition}} & \textbf{Batch} & \textbf{\tabincell{c}{Cache\\Enabled}} & \textbf{Execution model} & \textbf{\tabincell{c}{Communication\\protocols}}         & \textbf{\tabincell{c}{Open\\source}} \\ \hline
NeuGraph~\cite{neugraph_atc_2019}         & \multirow{7}{*}{Multi-GPU}    & N/A                          & 2D                                                                  & Row-wise                            & full-graph     & N                                      & Chunk-based              & Shared memory                                            & N                                    \\ \cline{1-1} \cline{3-10} 
PaGraph~\cite{pagraph_socc_2020}          &                               & Heuristics                   & LGD                                                                 & Shared graph store                  & k-hop          & Y                                      & -                        & N/A                                                      & Y                                    \\ \cline{1-1} \cline{3-10} 
BGL~\cite{bgl_nsdi23}                     &                               & Heuristics                   & LGD                                                                 & Shared graph store                  & k-hop          & Y                                      & Factored                 & N/A                                                      & N                                    \\ \cline{1-1} \cline{3-10} 
GNNLab~\cite{gnnlab}                      &                               & N/A                          & Shared graph store                                                  & Shared graph store                  & k-hop          & Y                                      & Factored                 & N/A                                                      & Y                                    \\ \cline{1-1} \cline{3-10} 
Rethinking~\cite{Rethinking_ics_2022}     &                               & N/A                          & Shared graph store                                                  & Row-wise                            & k-hop          & Y                                      & -                        & N/A                                           & N                                    \\ \cline{1-1} \cline{3-10} 
\blue{DSP~\cite{dsp-ppopp23}}             &                               & N/A                          & METIS                                                               & Row-wise                            & k-hop          & Y                                      & Operator-parallel        & N/A                                                      & Y                                    \\ \cline{1-1} \cline{3-10} 
\blue{Legion~\cite{legion_atc23}}         &                               & N/A                          & METIS                                                               & Row-wise                            & k-hop          & Y                                      & -                        & N/A                                                      & N                                    \\ \hline
ROC~\cite{roc_mlsys_2020}                 & \multirow{11}{*}{GPU-cluster} & Learning-based               & Edge-cut                                                            & Shared graph store                  & full-graph     & Y                                      & -                        & Shared memory                                            & Y                                    \\ \cline{1-1} \cline{3-10} 
P3~\cite{p3_osdi_21}                      &                               & N/A                          & Hash                                                                & Column-wise                         & k-hop          & Y                                      & Pull-push, Async.        & Weight staleness                                         & N                                    \\ \cline{1-1} \cline{3-10} 
Sancus~\cite{sancus_vldb_2022}            &                               & N/A                          & 1D                                                                  & Row-wise                            & full-graph     & Y                                      & Async                    & Embedding staleness                                      & Y                                    \\ \cline{1-1} \cline{3-10} 
NeutronStar~\cite{neugraph_atc_2019}      &                               & N/A                          & Edge-cut                                                            & Row-wise                            & full-graph     & Y                                      & -                        & P2P                                                      & Y                                    \\ \cline{1-1} \cline{3-10} 
DGCL~\cite{dgcl_eurosys_2021}             &                               & N/A                          & METIS                                                               & Row-wise                            & full-graph     & N                                      & -                        & P2P                                                      & Y                                    \\ \cline{1-1} \cline{3-10} 
PipeGCN~\cite{wan2022pipegcn}             &                               & N/A                          & METIS                                                               & Row-wise                            & full-graph     & Y                                      & Async.                   & \tabincell{l}{Embedding and\\gradient staleness}         & Y                                    \\ \cline{1-1} \cline{3-10} 
BNS-GCN~\cite{BNSGCN}                     &                               & N/A                          & Edge-cut                                                            & Row-wise                            & full-graph     & N                                      & -                        & P2P                                                      & Y                                    \\ \cline{1-1} \cline{3-10} 
DistDGLv2~\cite{distdglv2_kdd_2022}       &                               & Heuristics                   & METIS                                                               & Row-wise                            & subgraph       & Y                                      & Factored                 & N/A                                                      & N                                    \\ \cline{1-1} \cline{3-10} 
\blue{SALIENT++~\cite{salient++-mlsys23}} &                               & N/A                          & Edge-cut                                                            & Row-wise                            & full-graph     & Y                                      & -                        & N/A                                                      & Y                                    \\ \cline{1-1} \cline{3-10} 
\blue{G3~\cite{g3-sigmod23}}              &                               & Heuristics                   & Edge-cut                                                            & Row-wise                            & full-graph     & N                                      & -                        & P2P                                                      & N                                    \\ \cline{1-1} \cline{3-10} 
\blue{AdaQP~\cite{adaqp_mlsys23}}         &                               & N/A                          & Edge-cut                                                            & Row-wise                            & full-graph     & N                                      & -                        & Broadcast                                                & Y                                    \\ \hline
AliGraph~\cite{aligraph_vldb_2019}        & \multirow{8}{*}{CPU-cluster}  & N/A                          & \begin{tabular}[c]{@{}l@{}}Vertex-cut \\      Edge-cut\end{tabular} & Row-wise                            & k-hop          & Y                                      & -                        & N/A                                                      & Y                                    \\ \cline{1-1} \cline{3-10} 
AGL~\cite{agl_vldb_2020}                  &                               & N/A                          & DFS                                                                 & Row-wise                            & k-hop          & N                                      & -                        & N/A                                                      & N                                    \\ \cline{1-1} \cline{3-10} 
DistDGL~\cite{distdgl_ai3_2020}           &                               & Heuristics                   & METIS                                                               & Row-wise                            & k-hop          & Y                                      & -                        & N/A                                                      & Y                                    \\ \cline{1-1} \cline{3-10} 
CM-GCN~\cite{distgnn_CMGCN}               &                               & Operator-based               & Edge-cut                                                            & Row-wise                            & subgraph       & N                                      & Async.                   & Vertex-level                                             & N                                    \\ \cline{1-1} \cline{3-10} 
DistGNN~\cite{distgnn_sc_2021}            &                               & N/A                          & 2D                                                                  & Row-wise                            & full-graph     & Y                                      & Chunk-based, Async.      & \tabincell{l}{P2P, Pipeline-based,\\Embedding-staleness} & N                                    \\ \cline{1-1} \cline{3-10} 
FlexGraph~\cite{flexgraph_eurosys_2021}   &                               & Learning-based               & Par2E                                                               & Row-wise                            & full-graph     & Y                                      & Chunk-based              & Pipeline-based                                                 & N                                    \\ \cline{1-1} \cline{3-10} 
ByteGNN~\cite{bytegnn_vldb_2022}          &                               & Heuristics                   & LGD                                                                 & Row-wise                            & k-hop          & N                                      & Operator-parallel        & N/A                                                      & N                                    \\ \cline{1-1} \cline{3-10} 
\blue{ParallelGCN~\cite{spmm-vldb23}}     &                               & Heuristics                   & Hypergraph                                                          & Row-wise                            & full-graph     & N                                      & Chunk-based              & P2P                                                      & Y                                    \\ \hline
Dorylus~\cite{dorylus_osdi_2021}          & Serverless                    & N/A                          & Edge-cut                                                            & Row-wise                            & full-graph     & Y                                      & Async.                   & \tabincell{l}{Embedding and\\gradient staleness}         & Y                                    \\ \hline
SUGAR~\cite{sugar}                        & Low-resrouce devices          & N/A                          & Edge-cut                                                            & Row-wise                            & subgraph       & N                                      & -                        & N/A                                                      & N                                    \\ \hline
\end{tabular}
}
\vspace{-15pt}
\end{table*}


\blue{In recent years, many distributed GNN systems are proposed as a combination of distributed graph processing systems and distributed deep learning systems. The distributed GNN systems take the general idea of efficiently processing graph data from the former systems, while combining the capability of processing structured data from the latter systems. A recent survey systematically introduced the evolution from the graph processing system to the GNN system~\cite{distgnn-evolution-comp-survey23}.
Based on the taxonomy proposed in this survey,
we review \numsys distributed GNN training systems and frameworks as the representatives summarized in Table~\ref{tab:systems}, } 
and mainly classify them into three categories according to the computation resources: a) Systems on a single machine with multi-GPUs, b) Systems on GPU clusters and c) Systems on CPU clusters. Besides, there are a few distributed GNN training systems using more diversified computation resources, such as serverless threads, mobile devices, and edge devices. We classify them as miscellaneous systems. 
In the table, we make a summarization of the techniques the systems incorporate according to our new taxonomy. 
\blue{Note that for the column of execution model, we ignore the conventional, one-shot, and synchronous execution models as they are the basic execution models adopted by most systems. Therefore, we only note the execution models of systems with special designs.
For detailed introduction of each system and their relations, please refer to Section~\ref{sec:system-app} in the Appendix.
}
\section{Future Direction}
\blue{As a general solution to train on large-scale graphs, distributed GNN training gain widespread attention in recent years. In addition to the techniques and systems discussed above, there are other interesting but emerging research topics in distributed GNN training. 
We discuss some interesting directions in the following.
}

\textbf{Benchmark for distributed GNN training.} 
Many efforts are made in benchmarking traditional deep learning models. For instance, DAWNBench~\cite{coleman2017dawnbench} provides a standard evaluation criterion to compare different training algorithms. It focuses on the end-to-end training time consumed to converge to a state-of-the-art accuracy, as well as the inference time. Both single machine and distributed computing scenarios are considered. Furthermore, many benchmark suites are developed for traditional DNN training to profile and characterize the computation~\cite{mlperf_mlsys20,Zhu2018a,Dong2017b}. As for GNNs, Dwivedi~\cite{Dwivedi2020BenchmarkingGN} attempts to benchmark the GNN pipeline and compare the performance of different GNN models on medium-size datasets. 
Meanwhile, larger datasets~\cite{ogb_benchmark,graphgt_neurips2021,largedatabase_2021} for graph machine learning tasks have been published.
However, to our knowledge, few efforts have been made to compare the efficiency of different GNN training algorithms, especially in distributed computing scenarios.
GNNMark~\cite{gnnmark_ispass21} is the first benchmark suite developed specifically for GNN training. It leverages the NVIDIA nvprof profiler~\cite{nvprof_2012} to characterize kernel-level computations, NVBit framework~\cite{nvbit_2019} profiles memory divergence and further modifies PyTorch source code to collect data sparsity. However, it lacks flexibility for quick profiling of different GNN models and does not pay much attention to profiling the distributed and multi-machine training scenarios. 
Therefore, it would be practical for a new benchmark to be designed for large-scale distributed GNN training.

\textbf{Large-scale dynamic graph neural networks.}
In many applications, graphs are not static. The vertex attributes or graph structures often evolve with changes, which requires the representations to be updated in time. 
Li and Chen~\cite{li2021cache} proposed a general cache-based GNN system to accelerate the representation updating. It sets a cache for hidden representations and selects valuable representations to save time for updating.
DynaGraph~\cite{guan2022dynagraph} efficiently trains GNN via cached message passing and timestep fusion. Furthermore, it optimizes the graph partitioning in order to train GNN in a data-parallel method.
Although dynamic GNN has long been an interesting area of research, as far as we know, there are no more works that specifically focus on distributed dynamic GNN training. The dynamism, including features and structure, poses new challenges to the ordinary solution in a distributed environment. Graph partitioning has to quickly adjust to the change of vertices and edges while meeting the requirements of load balance and communication reduction. The update of the graph structure drops the cache hit ratio, which significantly influences the end-to-end performance of GNN training.

\textbf{Large-scale heterogeneous graph neural networks.} 
Many heterogeneous GNN structures are proposed in recent years~\cite{zhang2019heterogeneous, fu2020magnn, wang2019heterogeneous, zhao2021heterogeneous, chang2022megnn}. However, few distributed systems take the unique characteristic of heterogeneous graphs into consideration to support heterogeneous GNNs. Since the size of the feature and the number of neighbors may vary greatly for vertices with different types, processing heterogeneous graphs in a distributed manner may cause severe problems such as load imbalance and high network overhead. Paddle Graph Learning~\footnote{https://github.com/PaddlePaddle/PGL} framework provides easy and fast programming of the message passing-based graph learning on heterogeneous graphs with distributed computing. DistDGLv2~\cite{distdglv2_kdd_2022} takes the imbalanced workload partition into consideration and leverages the multi-constraint technique in METIS to mitigate this problem. More attention should be paid to this research topic to address the above problems.





\textbf{GNN model compression technique.}
Although model compression, including pruning, quantization, weight sharing, etc., is widely used in deep learning, it has not been extensively applied in distributed GNN training. 
The compression on network structures like pruning~\cite{gnncp_vldb_2021} can be contacted with a graph sampling strategy to solve the out-of-memory problem. Model quantization is another promising approach to improving the scalability of GNN models. SGQuant~\cite{sgquant_ictai_2020} is a GNN-tailored quantization algorithm and develops multi-granularity quantization and auto-bit selection. Degree-Quant~\cite{degreequant_iclr_2021} stochastically protects (high-degree) vertices from quantization to improve weight update accuracy. BinaryGNN~\cite{Binary_CVPR_2021} applies a binarization strategy inspired by the latest
developments in binary neural networks for images and knowledge distillation for graph networks. 
For the distributed settings, 
recently, Song et al.\cite{song2022ec} proposed EC-Graph for distributed GNN training with CPU clusters which aims to reduce the communication costs by message compression. It adopts lossy compression and designs compensation methods to mitigate the induced errors. Meanwhile, A Bit-Tuner is used to keep the balance between model accuracy and message size. GNN model compression is orthogonal to the aforementioned distributed GNN optimization techniques, and it deserves more attention to help improve the efficiency of distributed GNN training.



\section{Conclusion}

Distributed GNN training is one of the successful approaches of scaling GNN models to large graphs. 
In this survey, we systematically reviewed the existing distributed GNN training techniques from graph data processing to distributed model execution and covered the life-cycle of end-to-end distributed GNN training. 
\blue{We divide the distributed GNN training pipeline into three stages, data partition, batch generation, and GNN model training, 
which heavily influence the GNN training efficiency. 
To clearly organize the new technical contributions which optimize these stages, we proposed a new taxonomy that consists of four orthogonal categories
-- GNN data partition, GNN batch generation, GNN execution model and GNN communication protocol.}
In the GNN data partition category, we described the data partition techniques for distributed GNN training;
in the GNN batch generation category, we presented the techniques of fast GNN batch generation for mini-batch
distributed GNN training; 
\blue{in the GNN execution model category, we discussed the execution model used in mini-batch and full-graph training respectively;} 
in the GNN communication protocol category, we discussed both synchronous and asynchronous protocols for distributed GNN training.
After carefully reviewing the techniques in the four categories, we summarized existing representative distributed GNN systems for multi-GPUs, GPU-clusters and CPU-clusters, respectively, 
\blue{and gave a discussion about the future direction in optimizing distributed GNN training.}


\bibliographystyle{ACM-Reference-Format}
\bibliography{main.bib}

\clearpage
\appendix

\section{Distributed GNN Training Systems}
\label{sec:system-app}

\begin{figure}[H]
    \centering
    \resizebox{\textwidth}{!}{
        \includegraphics{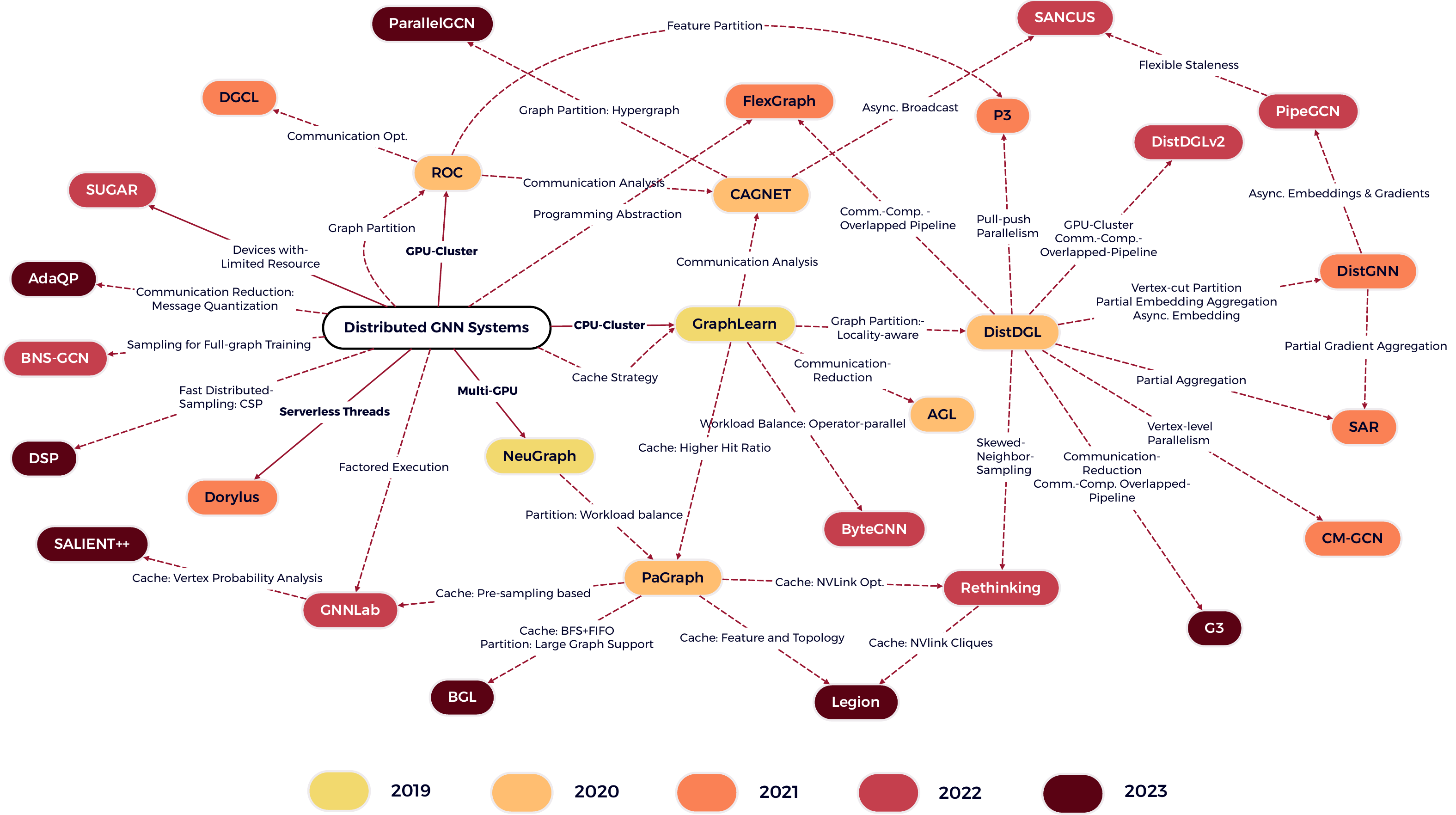}
    }
    \caption{The evolution of distributed GNN systems.}
    \label{fig:system-evolution}
\end{figure}

\blue{In this section, we present detailed introduction of the representative systems presented in Table~\ref{tab:systems}. 
We systematically summarize the evolution of these systems in Figure~\ref{fig:system-evolution}. The evolution of techniques are marked on the edges.}
In the following, we highlight the main techniques leveraged by each system to optimize the training efficiency and further introduce some other system settings and optimizations not presented in the table.

\subsection{Systems on single machine with multi-GPUs}
GPUs are the most powerful computing resource to train neural network models. A common solution for distributed GNN training is to use multi-GPUs on a single machine. Here we introduce some popular systems that support efficient multi-GPUs training. 


\textbf{NeuGraph}~\cite{neugraph_atc_2019} is a pioneer work that aims at enabling scalable full-graph GNN training by bridging deep learning systems and graph processing systems with a novel programming abstraction called SAGA-NN. 
To support the execution of the SAGA-NN model on large graphs, it adopts a 2D graph partition based on locality-aware algorithms, which partitions the large graph into disjoint edge chunks. 
These edge chunks are scheduled sequentially to each GPU to perform chunk-based computation graph execution.
In order to better improve the performance of CPU-GPU data transfer, a selective scheduling strategy is proposed, which filters useful vertices only from a large chunk and sends them to GPU. 
A chain-based chunk scheduling scheme is designed to fully utilize the PCIe bandwidth, and maximize the performance in multi-GPU GNN training.
Other optimizations are designed to optimize the computation in GPU kernels, such as redundant computation elimination and stage fusion, by considering both the character of graph structure and specific GNN models. 
Overall, NeuGraph, implemented on top of TensorFlow with C++ and Python, naturally combines deep learning systems with graph computation and fully utilizes the efficiency of the CPU's large memory, PCIe's hierarchical structure, and GPU's computing power.
\textbf{PaGraph}~\cite{pagraph_socc_2020} is a novel GNN framework that was implemented under DGL and PyTorch, supporting sampling-based GNN training on multi-GPUs. 
The basic idea of PaGraph is to leverage the spare GPU memory and use it as a cache during sampling-based GNN training. 
It mainly focuses on the workload imbalance and redundant vertex access problem, and designs a heuristic partitioning strategy, by considering the affinity of each vertex to a partition. 
To support multi-GPUs and reduce the storage overhead of partitions, PaGraph removes the redundant vertices and edges that have no contribution to GNN training after the partitioning. 
To further achieve computing efficiency, resource isolation is conducted to avoid interference between different operations, and local shuffling is added to improve data parallel training.
To sum up, PaGraph accelerates GNN training speed by cache and graph partition strategies and eventually achieves up to 96.8\% reduction of data transfer and up to 16$\times$ performance improvement compared to DGL. 

{\textbf{BGL}~\cite{bgl_nsdi23} is a GPU-efficient GNN training system for large graph learning, which focuses on sampling-based GNN. 
To remove heavy network contention of feature retrieving, BGL adopts a two-level cache consisting of multi-GPU and CPU memory to maximize cache size. After partitioning the graph via BFS and assignment heuristic, in order to improve training speed, BGL further optimizes resource allocation via profiling-based resource allocation. The computation in BGL is split into eight stages, and the profiling-based resource allocation method distributes isolated resources to different stages with the goal of minimizing the maximal completion time of all stages. In conclusion, BGL is a generic system that can be applied to various computation backends (DGL, Euler\footnote{https://github.com/alibaba/euler}, PyG) and pushes GPU utilization to a significantly higher level compared to existing frameworks.}


{\textbf{GNNLab}~\cite{gnnlab} is a newly GNN system optimized for sampling-based GNN models and conducts an overall investigation on a conventional designed sampling-based system. It originally introduces a cache policy based on pre-sampling, which achieves efficient and robust results on various sampling algorithms and datasets. Based on the {factored design for sampling-based training}, GNNLab further solves the challenge of workload imbalance across GPUs by assigning GPUs to different executors on demand of workload. GNNLab first assigns a certain number of GPUs calculated by the heuristic equation. And then, in order to choose the optimal GPU allocation constantly, GNNLab dynamically switches executors according to standby and existing Trainer working efficiency. In summary, this system eliminates heavy data I/O on feature retrieving and GPU memory contention of conventional GNN execution model, outperforming DGL and PyG by up to 9.1$\times$ and 74.3$\times$}

{\textbf{Rethinking}~\cite{Rethinking_ics_2022} develops a data-parallel GNN training system, focusing on the data movement optimization among CPU and GPUs. Unlike other systems, it allows GPUs to fetch data from each other by leveraging the high bandwidth among them. It builds a cost model to estimate the cost of data movement among CPU and GPUs, which can be minimized by the optimal data placement strategy. Based on the cost model, it formulates the data placement problem as a constrained optimization problem and develops an efficient solution to find an excellent data partition. In addition, to further reduce the data movement, it introduces a \emph{locality-aware neighbor sampling} to increase the sampling probability of vertices on GPU which have little loading time. As a result, empirical studies demonstrate that it achieves better data loading efficiency than DGL~\cite{dgl_arxiv_2019} and PaGraph ~\cite{pagraph_socc_2020} with the smart data placement strategy.}

\blue{
\textbf{DSP}~\cite{dsp-ppopp23} introduces a distributed sampling technique to accelerate the batch generation stage in distributed GNN training. It presents a \textit{Collective Sampling Primitive} (CSP) that sends sampling tasks to remote workers, receiving only the sampled results instead of the entire neighbor list. This reduces communication costs as the sampled result requires fewer bits compared to the complete neighbor list. To support CSP effectively, DSP prioritizes caching the partitioned graph topology on each GPU. Theoretical analysis confirms that CSP is versatile in expressing both node-wise sampling and layer-wise sampling for GNN batch generation.
Moreover, DSP implements a mini-batch pipeline, assigning a sampler, a loader, and a trainer for each GPU. A carefully designed centralized communication coordination prevents deadlocks in the pipeline. The distributed sampling technique proposed by DSP achieves up to 20$\times$ speedup compared to existing CPU-based and UVA-based sampling methods. In most cases, the overall training time is improved by over $2\times$ compared to DGL~\cite{dgl_arxiv_2019}.
}

\blue{
\textbf{Legion}\cite{legion_atc23} focuses on multi-GPU training with NVLink connections. It introduces a hierarchical graph partitioning mechanism that considers the structure of NVLink cliques. It first partitions the graph with METIS, and assign each partition to an NVLink Clique. Utilizing the fact that communication among GPUs in the same NVLink clique is more cost-effective than PCIe-connected GPUs, Legion sets up a public cache for GPUs in each NVLink Clique, which is stored distributively in that NVLink Clique. 
Legion employs the batch generation mechanism similar to PaGraph~\cite{pagraph_socc_2020}, by only sampling the training vertices from a partition assigned to an NVLink clique. Therefore, each graph partition enjoys a dedicated cache, thus improving the cache hit ratio. 
Inspired by GNNLab~\cite{gnnlab}, Legion also utilizes a pre-sampling mechanism to identify the most frequently accessed vertices for graph sampling and feature extraction. Legion proposes a caching strategy that caches both the graph topology and vertex features, which is also proposed in a concurrent work~\cite{ducati_sigmod23}. The ratio of GPU memory allocation for topology cache and feature cache is determined by a parameter $\alpha$. A linear search is performed to find the optimal $\alpha$, by carefully modeling the cache benefits with a given $\alpha$. Legion successfully pushes the envelope for multi-GPU GNN training to billion-scale graphs and achieves significant speedup for end-to-end GNN training. 
}

\subsection{Systems on GPU clusters}
GPU cluster is an extension of a single machine with multi-GPUs, and it provides multi-machine multi-GPUs for the case when graphs and input features are too large to fit in a single machine. With the GPU cluster, both the communication between CPU and GPU and the communication among machines should be considered.
The following systems have been tested on GPU clusters according to their original experiments.


\textbf{ROC}~\cite{roc_mlsys_2020} is a distributed full-graph GNN training and inference system which is implemented on top of a GPU-optimized graph processing system {FlexFlow~\cite{MLSYS2019_c74d97b0}}. ROC supports both multi-GPUs and GPU clusters. In ROC, GPU memory is used as a cache and CPU memory is to store all GNN tensors and intermediate results that support larger GNN. So, ROC only requires all GNN tensors to fit in the CPU DRAM and each GPU access the graph and features not at local via data transfers between CPU and GPU. ROC applies a {learning-based graph partitioner} to balance the workload and applies a dynamic programming-based memory manager to minimize data transfers between CPU and CPU memories. The memory manager is guided by the topology of the GNN architecture and determines the tensors to be cached in GPU by analyzing GNN's future operations. 

{\boldsymbol{$P^3$}~\cite{p3_osdi_21} is a distributed system that introduces model parallel into GNN training, and orchestrates it with the data parallelism to reduce the total communication cost. 
$P^3$ takes the observation that a large communication overhead is introduced in transferring the vertex features in the first layer, and designs a column-wise partitioning strategy to distribute the features to different workers. Specifically, different from traditional data-parallel GNN training, $P^3$ partitions the graph structure and vertex features independently. During the training process, in the first layer, each worker handles the given columns of features of all vertices and fetches graph structures from other workers. In such way, only the graph structure is transferred among workers, which incurs less network overhead than transferring the vertex features. 
For the remaining layers, traditional data parallelism is adopted. This hybrid model-data parallelism is named \textit{push-pull parallelism} in $P^3$. Under push-pull parallelism, a simple hash partitioning of the graph structure is sufficient. Moreover, a mini-batch pipeline is designed to fully utilize the computing resources, and an asynchronous communication protocol with bounded staleness is used. $P^3$ firstly incorporates the model parallelism into distributed GNN training for a good cause and designs a new training approach to improve overall performance, especially when the feature size is large. 
}

{\textbf{SANCUS}~\cite{sancus_vldb_2022} is a decentralized GNN training framework that introduces historical embedding in full graph training, in order to reduce the large communication overhead by broadcasting the latest embeddings.
Unlike traditional parameter server-based architecture, it applies decentralized training architecture to eliminate the centralized bottleneck bandwidth. 
The graph and features are co-partitioned among a set of workers (e.g., GPUs). 
To avoid massive communication among workers, a new communication primitive, called skip-broadcast, is proposed to support staleness-aware training. 
Skip-broadcast is compatible with collective operations like ring-based pipeline broadcast and all-reduce, and it reshapes the underlying communication topology by automatically removing communications when the cached embedding is not too stale.
Three different staleness metrics are supported in SANCUS among which epoch-adaptive variation-gap embedding staleness works the best in general. 
Empirical results show that SANCUS reduces up to 74\% communications without losing accuracy.
}

{\textbf{NeutronStar}~\cite{neutronstar_sigmod_2022} is a distributed GNN training system with hybrid dependency management approach. It first introduces a cost model to estimate the redundant computation caused by storing duplicated vertices in different GPUs and the communication overhead caused by transferring boundary vertices. Further, it proposes a greedy-based heuristic algorithm to split dependencies into cached and communicated groups that can minimize the cost of accessing dependencies. In addition, NeutronStar decouples the dependency management from graph operation and NN functions for flexible auto differentiation and provides high-level python APIs for both forward and backward.
To sum up, NeutronStar combines two dependencies accessing mechanisms for GPU clusters GNN speedy training and integrates some optimizations for CPU-GPU heterogeneous computation and communication, outperforming DistDGL and ROC by up to 1.8$\times$ and 14.3$\times$}.

\textbf{DGCL}~\cite{dgcl_eurosys_2021} is a GNN-oriented communication library for GPU clusters. It takes the physical communication topology and the communication relation determined by the GNN model into consideration for planning the communication. {On the basis of the observation that hierarchies are always existing in the communication topology, DGCL splits the data transfer into stages, which are corresponding to the number of links between the source GPU and destination GPU. 
Thereafter, DGCL introduces a heuristic cost model for each communication plan.} And it obtains the optimal communication plan via an efficient tree spanning algorithm. As regards implementation, DGCL utilizes MPI and PyTorch to support the distributed environment and applies Horovod and PyTorch to support distributed model synchronization. 
According to the authors' empirical studies, this library reduces the communication time of peer-to-peer communication by 77.5\% on average and the training time for an epoch by up to 47\%.


{\textbf{PipeGCN}~\cite{wan2022pipegcn} is a full-graph distributed training system developed on top of DistDGL that introduces stale embeddings to form a pipeline. For neighbors on remote workers, PipeGCN leverages a deferred communication strategy to use both the historical embeddings and gradients in the previous epoch. The latest embeddings are still computed for local neighbors. 
With a mixture of fresh and stale features (gradients), PipeGCN can better overlap communication overhead with computation cost. 
Further, PipeGCN gives detailed proof that guarantees its bounded error and convergence. 
In addition, to reduce the effect of stale information, PipeGCN proposes a light-weight moving average to smooth the fluctuations of boundary vertices' features (gradients). 
PipeGCN is a representative system equipped with an asynchronous computation graph execution model and asynchronous communication protocol. With a simple strategy to fix the staleness bound to 1 epoch, PipeGCN is able to reduce the end-to-end training time by around 50\% without compromising the accuracy.
}

{\textbf{BNS-GCN}~\cite{BNSGCN} aims to find a balance between the large communication overhead of the full-graph training method and the accuracy loss of sampling-based training methods. The idea is to communicate only a portion of the boundary vertices in each remote worker. It reduces the number of transferred boundary vertices randomly by a factor of $\frac{1}{p}$.
Each partition selects the vertices from the boundary set stochastically and merely uses their features for GNN training. 
Thanks to its simple strategy, BNS-GCN can be merged into any distributed GNN training system based on edge-cut partitioning without introducing excessive computation. 
Implemented on top of DGL, this strategy achieves full-graph accuracy while the sampling method speeds up the training time. This simple combination of the two basic GNN acceleration techniques, i.e. the distributed GNN and sampling-based GNN, leads to an interesting training paradigm for further study. 
}

{\textbf{DistDGLv2}~\cite{distdglv2_kdd_2022} is a continuous work of DistDGL by the DGL team from Amazon.
It takes the characteristic of the heterogeneous graph into consideration for balanced graph partition.  
Specifically, it extends METIS to split heterogeneous graphs regardless of their vertex types and adopts hierarchical partitioning to first reduce the communication across the network and then reduce the data copy to GPUs in a mini-batch. 
In addition, it designs a mini-batch generation pipeline, to enable parallel computation of several mini-batches in the CPU.
A distributed hybrid CPU/GPU sampling technique is designed to fully utilize the GPU computation, which samples vertices and edges for each hop of the neighborhood in CPU and performs graph compaction in GPU to remove unnecessary vertices and edges for mini-batch computation. 
This mini-batch pipeline proposed achieved 2-3$\times$ total training speed up compared to DistDGL, and will soon be merged and released in the DGL source code. 
}

\blue{
\textbf{SALIENT++}~\cite{salient++-mlsys23} extends SALIENT to distributed GNN training and optimizes the communication cost by designing an analysis-driven cache strategy. It designs a propagation model to compute the probability of each vertex being sampled. Given a GNN model with specific sampling strategy, the sampled probability of each neighbor vertex of a center vertex can be accurately computed. With the propagation model, given graph partition and the initial probability of each training vertex being sampled into a training batch, the sampled probability of each vertex in the partition is computed. SALIENT++ use this probability to select important vertices into the cache for each worker. Furthermore, SALIENT++ proposes a fine-grained pipeline composed of 10 stages for feature collection. Experiments show that SALIENT++ achieves a $12.7\times$ speedup compared to DistDGL~\cite{distdgl_ai3_2020}.
}

\blue{
\textbf{G3}~\cite{g3-sigmod23} focuses on the scalable full-graph training on GPU-clusters. Graph partitioning and fine-grained pipeline are the two main concerns in G3. First, G3 designs a two-step graph partitioning strategy to balance the workload and the communication cost on each worker. The first step employs the multi-constraint METIS partition, setting the balance of the number of vertices and number of edges in each partition as the two constraints, while minimizing the total communication cost which is modeled by the number of cut edges. In the second step, G3 iteratively exchanges vertices between partitions with the largest and smallest number of remote neighbors. The second step ensures a communication-balanced workload partition, by roughly making each partition involving a similar number of remote neighbors. Second, G3 proposes a fine-grained pipeline for full-graph distributed training. A bin-packing mechanism is designed, which packs the training vertices in one GPU into different bins with the help of the scheduling algorithm proposed. On the local worker, the execution of a GNN layer is split into the execution of bins, which can be overlapped. On a remote worker, some vertices may be ready to process the next layer once all their remote neighbors' embeddings are received. Therefore, these vertices are packed into a bin. The execution of the next GNN layer of vertices in this bin starts immediately. This fine-grained pipeline implements both intra-layer pipeline and inter-layer pipeline, fully overlapping the communication with computation. G3 is the first distributed system that proposes a well-designed pipeline in full-graph synchronous GNN training. 
}

\blue{
\textbf{AdaQP}~\cite{adaqp_mlsys23} aims at reducing the communication overhead among workers in full-graph distributed GNN training. AdaQP proposes an innovative approach for distributed full-graph training using stochastic quantization of messages to reduce communication traffic and enhance efficiency. AdaQP quantizes messages into lower-precision integers before transmission across devices. An adaptive bit-width assignment scheme is proposed to optimize quantization bit-widths to balance data volume distribution between devices, leading to improved training efficiency. Theoretical analysis validates fast training convergence at $\mathcal{O}(T^{-1})$ for $T$ total training epochs. Furthermore, AdaQP also proposes a parallelization mechanism on each worker, by executing the computation of vertices without remote neighbors (\textit{central nodes}) in advance. Simultaneously, the remote neighbors of other vertices (\textit{marginal nodes}) are being transferred. Therefore, the computation of central nodes and the communication of neighbors of marginal nodes are overlapped. This overlapping insight is also brought up by G3~\cite{g3-sigmod23}. Experiments show that the message quantization improves the throughput of GNN training by up to $3.02\times$ yet incurs a negligible accuracy drop. 
}


\subsection{Systems on CPU clusters}
Compared to the GPU cluster, the CPU cluster is easy to scalable and economical. Existing big data processing infrastructure is usually built with CPU clusters. There are many industrial-level distributed GNN training systems running on CPU clusters. 

{\textbf{AliGraph}~\cite{aligraph_vldb_2019} is the first industrial-level distributed GNN platform developed by Alibaba Group. 
The platform consists of the application, algorithm, storage, sampling, and operator layers, where the latter three layers form the GNN system. 
The storage stores and organizes the raw data to support fast data access. 
It applies various {graph partition} algorithms to adapt to different scenarios, stores the attributes separately to save space and uses a {cache to store important vertices} to reduce communication. 
The sampling layer abstracts three kinds of sampling methods and the operator layer abstracts the GNN computation with two kinds of operators.
Users can define their own models by choosing their sampling methods and implementing their own operators. 
Therefore, AliGraph supports not only the running of in-house developed GNNs and classical graph embedding models, but also the quick implementation of the latest SOTA GNN models.
}

{\textbf{AGL}~\cite{agl_vldb_2020} is a distributed GNN system, with fully-functional training and inference designed for industrial purpose graph machine learning. AGL follows the message passing scheme and mainly supports sampling-based GNN training. The system consists of three major modules -- GraphFlat, GraphTrainer, and GraphInfer. GraphFlat is a distributed $k$-hop neighborhood batch generator implemented with MapReduce. It merges in-degree neighbors and propagates merged information to out-degree neighbors via message passing. Besides, it follows a workflow of sampling and indexing to eliminate the adverse effects caused by graph skewness. GraphTrainer is the distributed training framework following the parameter server. It leverages pipeline, pruning, and edge-partition to eliminate the overhead on data I/O and optimize floating point calculations. GraphInfer is a distributed inference module that splits $K$ layer GNN models into $K$ slices and applies message passing $K$ times based on MapReduce. This schema can eliminate redundant computations and reduce inference time costs dramatically. In summary, AGL achieves a nearly linear speedup in training with 100 workers, being able to finish a 2-layer GAT training with billions of vertices in acceptable hours.}

{\textbf{DistDGL}~\cite{distdgl_ai3_2020} is a distributed version of DGL designed by Amazon, which performs efficient and scalable mini-batch GNN training on the CPU cluster. 
The main components in DistDGL are the distributed samplers, distributed Key-Value (KV) store, trainers, and a dense model update component. 
The samplers are in charge of mini-batch generation and expose a set of flexible APIs to help the user define various sampling algorithms. 
The KVStore stores the graph and features distributedly and supports separate partition methods for vertex data and edge data. 
The trainers compute gradients of parameters over a mini-batch. 
While the dense model parameters are updated with synchronous SGD, the sparse vertex embeddings are updated with asynchronous SGD.
As one of the earliest open-source systems and frameworks for distributed GNN training, DistDGL has made a great impact on the fast implementation of ideas for new training algorithms or distributed systems.  
}

{\textbf{CM-GCN}~\cite{distgnn_CMGCN} proposes a new training model based on the characteristic of distributed graph partitioning and mini-batch selection. Its basic idea is to form cohesive mini-batches, in which training vertices are connected tightly and share common neighbors, in order to reduce the communication cost in retrieving neighbor vertices from remote workers. A well-designed cost model is proposed to partition the workload, which we classify as an operator-based model. 
In addition, CM-GCN enables a vertex-level asynchronous computation by decomposing the computation of vertices and processing them based on their data availability. 
Similar to BNS-GCN, CM-GCN introduces a new GCN training paradigm that attempts to reduce communication by focusing less on remote data. 
Although this idea may introduce training bias compared with the original GCN, experiment results show that both BNS-GCN and CM-GCN maintain a similar accuracy compared with GCN. 
}

\textbf{DistGNN}~\cite{distgnn_sc_2021} also focuses on full graph GNN training and adds optimizations on top of DGL to support distributed GNN training. 
For distributed memory GNN training, DistGNN adopts \textit{Libra}~\cite{libra_nips22} fast partitioning with balanced workloads. {By partitioning the graph in a vertex-cut manner}, DistGNN designs three different aggregation strategies, including bias aggregation (by ignoring remote neighborhood), synchronous aggregation, and asynchronous aggregation with bounded staleness. The asynchronous aggregation adopts the Type I asynchronous execution model and effectively reduces communication costs. 
DistGNN is implemented on top of DGL with an optimized backend with C++ and LIBXSMM library. With single socket SPMM and distributed Libra partitioning being merged to the DGL source code, DistGNN has made a great contribution to the open-source GNN training.  


{\textbf{FlexGraph}~\cite{flexgraph_eurosys_2021} is a distributed full-graph GNN training framework with a new programming abstraction NAU (a.k.a., NeighborSelection, Aggregation and Update). NAU is a flexible GNN framework to tackle the challenges that GAS-like GNN programming abstractions are unable to express GNN models using indirect neighbors during the aggregation phase. To support the new abstraction, FlexGraph employs hierarchical dependency graphs (HDGs) to compactly store the ``neighbors'' with different definitions and hierarchical aggregation strategies in GNN models. In addition, FlexGraph applies a hybrid aggregation strategy to distinguish the aggregation operations in different contexts and design suitable methods according to their characteristics. This system is implemented on top of PyTorch and libgrape-lite, a library for parallel graph processing. Overall, based on novel NAU abstraction, FlexGraph adopts various optimizations from different aspects, so that it gains better expression and capability of scaling than DGL, PyG, NeuGraph, and Euler.}

{\textbf{ByteGNN}~\cite{bytegnn_vldb_2022} is a distributed GNN training system designed by ByteDance Incorporation, focusing on improving resource utilization during mini-batch training. 
Noticing that GPU does not bring in too many benefits for the sampling process, ByteGNN focuses on fully utilizing the capability of CPU clusters.
An abstraction of the sampling phase for both supervised and unsupervised GNN models is proposed. 
With such abstraction, the workflow of the sampling phase can be {arranged as a DAG corresponding to the operator-parallel execution model}.
A two-level scheduling strategy is designed to schedule different DAGs and their inner operators. 
ByteGNN also attempts to reduce the large network overhead by addressing the workload imbalance problem and designing a new graph partitioning method. 
Implemented based on GraphLearn (i.e. AliGraph), ByteGNN significantly improves the CPU utilization and achieves up to 3.5$\times$ speed up compared with DistDGL.
}

\blue{
ParallelGCN~\cite{spmm-vldb23} focusing on accelerating the distributed SpMM execution of Graph Convolutional Network. The authors demonstrate that the broadcast communication protocol employed by CAGNET~\cite{gespmm_sc20} is inefficient since redundant embeddings of some vertices are transferred. To improve the efficiency of communication, ParallelGCN adopts P2P-based communication protocol, and only sends vertex embeddings that are necessary between workers. ParallelGCN proposes a non-blocking P2P communication operation, which implements the chunk-based execution model, by overlapping the computation of received chunks and communication of other chunks. ParallelGCN also focuses on existing graph partitioning strategies and points out that the hypergraph partition model is more expressive than the existing graph partition model for distributed GNN training. Thanks to the hypergraph partition model, ParallelGCN is able to model the communication cost between each worker pair more accurately and achieves up to $41\%$ communication reduction. Based on the 1D partitioning distributed GNN execution, ParallelGCN introduces optimizations in both the communication protocol and data partition strategies. 
}



\subsection{Miscellaneous}
\textbf{Dorylus}~\cite{dorylus_osdi_2021} studies the possibility of leveraging a more affordable computing resource, i.e., serverless threads, in GNN training. GNN computing is decomposed into different operations, including graph operations and neural operations. CPU clusters are used for graph operations such as feature extraction and feature aggregation, while the serverless threads are used for dense matrix multiplication in neural operations. Several optimizations for serverless threads management are specifically designed in Dorylus. Pipelining the GNN computation is another major contribution to Dorylus. 
SAGA-NN model is adopted in Dorylus, where different operators (e.g. Scatter, ApplyVertex) are executed in parallel, forming the execution pipeline.
The training vertices are split into different vertex intervals. 
Dorylus continuously feeds the training pipeline with vertex intervals, 
while different vertex intervals in the pipeline are executed by different operators at the same moment. A synchronization is introduced by such a mini-batch pipeline so that an asynchronous communication protocol with bounded staleness is designed. It is shown that the serverless threads achieve high efficiency close to GPUs yet are much cheaper, and the asynchronous execution pipeline still reaches higher accuracy compared with sampling-based GNN training methods. 


{\textbf{SUGAR}~\cite{sugar} is proposed to support resource-efficient GNN training. It uses multiple devices with limited resources (e.g., mobile and edge devices) to train GNN models and provides rigorous proof of complexity, error bound, and convergence about distributed GNN training and graph partition. After generating the subgraphs with {weighted graph and graph expansion}, each device in SUGAR trains local GNN models at the subgraph level. SUGAR maintains $K$ local models instead of a global model by keeping the local model updates within each device, such that the parallel training is able to save the computation, memory, and communication costs. In summary, SUGAR aims at enlarging the scalability of GNN training and achieves considerable runtime and memory usage compared to the SOTA GNN models on the large-scale graph.}






\end{document}